\documentclass{article}
\PassOptionsToPackage{table}{xcolor}

\usepackage{dpp_preprint,times}

\usepackage[T1]{fontenc}
\usepackage[utf8]{inputenc}
\usepackage{microtype}
\usepackage{amsmath,amssymb,bm,mathtools}
\usepackage{graphicx}
\usepackage{booktabs}
\usepackage{multirow}
\usepackage{xcolor}

\usepackage[section]{placeins}
\usepackage{titletoc}
\usepackage[hidelinks]{hyperref}
\definecolor{kaistblue}{RGB}{20, 135, 200}
\hypersetup{
    colorlinks=true,
    citecolor=kaistblue,
    linkcolor=kaistblue,
    urlcolor=kaistblue
}
\usepackage{url}

\definecolor{lightgray}{gray}{0.94}
\definecolor{darkgreen}{RGB}{32,112,72}

\title{Dynamic Manipulation with World--Action Models via Counterfactual Planning}
\author{%
\normalfont
\makebox[0.96\textwidth][c]{%
\begin{tabular}{c}
	\textbf{Sunwoo Park}$^{1,*}$ \quad \textbf{Wonbin Lee}$^{1,*}$ \quad \textbf{Seonghyun Jin}$^{1,*}$ \quad \textbf{Youngmin Kim}$^{1,*}$ \\
	\textbf{Jangho Park}$^{1}$ \quad \textbf{Jong Chul Ye}$^{1,\dagger}$ \\[0.55em]
$^1$Kim Jaechul Graduate School of AI, KAIST, Korea\\[0.2em]
{\small $^*$Equal contribution. $^\dagger$Corresponding author.}
\end{tabular}%
}}
\date{}

\usepackage{wrapfig}
\usepackage{enumitem}

\begin{document}

\maketitle
\fancyhead{}
\lhead{arXiv preprint}

\begin{abstract}
World--Action models (WAMs) trained on static demonstrations often fail to manipulate moving targets even when they possess the required manipulation skills. We attribute this failure to \textit{target-response collapse}: as execution advances, the policy becomes increasingly biased toward the learned continuation of its ongoing behavior and less responsive to target relocation.
To bridge the gap between what the model has learned and what it can generate from the current context, we formulate dynamic manipulation as counterfactual planning by \textit{decoupling the context used for plan generation from the physical state used for execution}. Our framework, \textbf{Dynamic Predictive Planning (DPP)}, first uses the WAM's predictive rollout to estimate when an interaction is expected to occur, and combines this timing estimate with observed target motion to predict the target's future interaction position. DPP then constructs a counterfactual observation that places this predicted target position in a familiar robot context, allowing the model to invoke an existing manipulation skill rather than generate a recovery behavior from an unfamiliar robot--target configuration. The resulting plan is connected to the robot's actual state during execution.
DPP enables real-time dynamic manipulation on a single consumer GPU without additional training on dynamic data. Experiments in simulation and on a real robot demonstrate consistent improvements across diverse target motions, with simulation performance surpassing all evaluated baselines, including methods additionally trained on dynamic data.
\textbf{Project~page:} \url{https://methoder00.github.io/DPP/}
\end{abstract}

\section{Introduction}

\begin{figure}[t]
    \centering
    \begingroup
    \graphicspath{{./}{figures/}}
    \includegraphics[width=\linewidth]{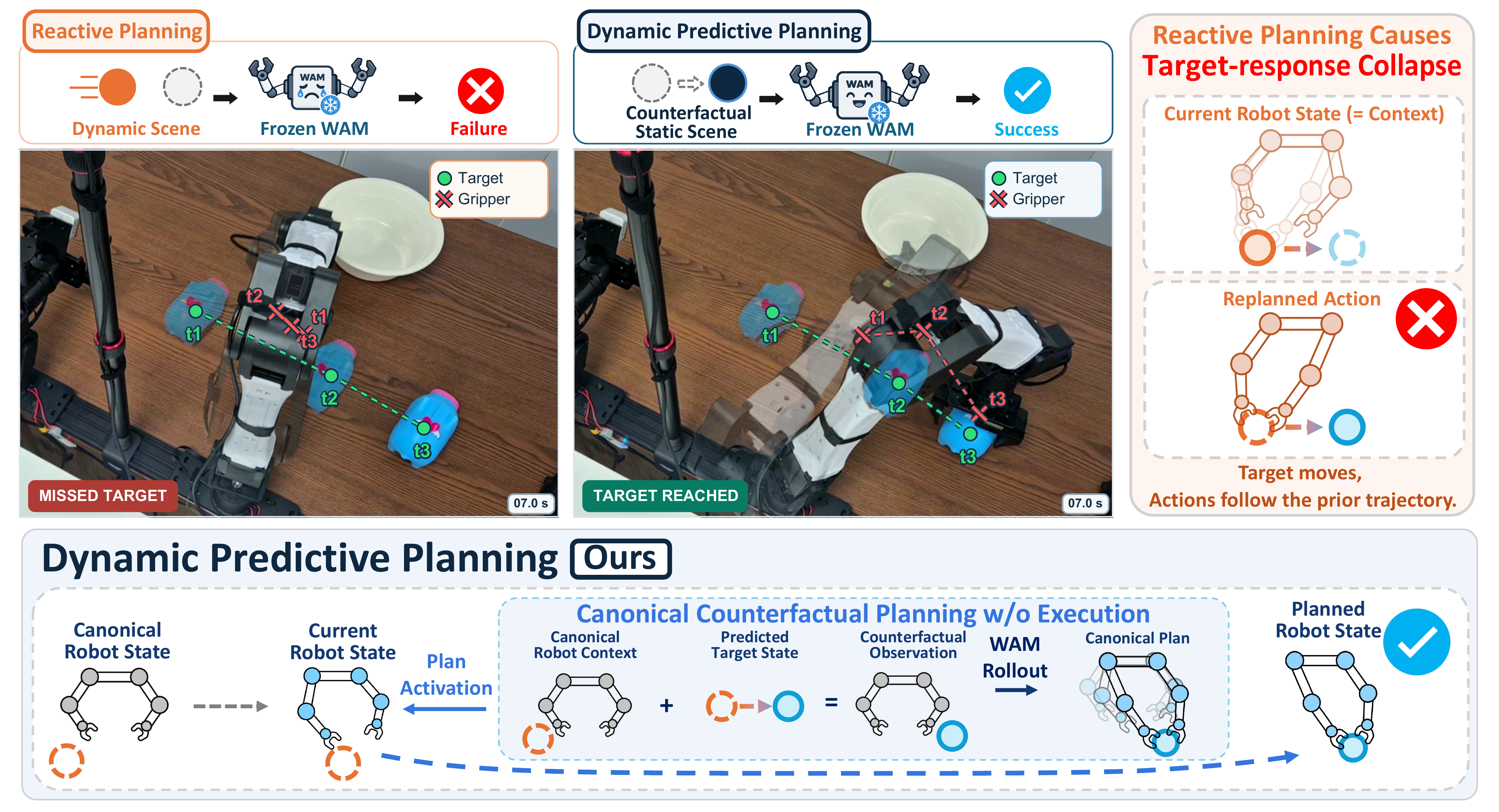}
    \endgroup    \caption{\textbf{Predict, don't chase.} DPP pairs the initial robot state with the predicted interaction position to invoke learned manipulation capabilities rather than repeatedly chasing the moving target.}
    \label{fig:predict-dont-chase}
    \vspace{-8pt} % Retain one line of separation below the figure.
\end{figure}

Vision--Language--Action (VLA) models map visual observations and language instructions directly to robot actions, achieving strong performance across diverse manipulation tasks~\citep{brohan2023rt,kim2024openvla,bjorck2025gr00t,intelligence2025pi_}. More recently, World--Action Models (WAMs) have augmented action generation with prediction of future visual states, enabling models to represent how scenes evolve as actions unfold~\citep{shen2026videovla,ye2026world,yuan2026fast,zhang2026imagewam,agarwal2026cosmos,LiL-RSS-26}.

In real-world deployment, however, robots must often interact with moving objects and respond to changes in the environment. Most manipulation skills are still learned from demonstrations collected with stationary targets and relatively canonical robot configurations~\citep{liu2023libero,mu2025robotwin}. As a result, a skill that is reliable in static settings may fail once the target moves during execution. This raises a central question: can such pretrained manipulation skills be reused in dynamic environments without collecting additional dynamic data or retraining the model?

A natural strategy is to repeatedly replan from the latest observation as the target moves. However, this can place the policy in robot--target configurations that are poorly supported by the static demonstrations: the robot has progressed to a later execution state, while the target has relocated to a position that was never observed together with that state during training. As execution progresses, the policy may therefore become increasingly biased toward the learned continuation of its current behavior, despite new visual evidence that the target has moved. It then continues toward a familiar subgoal instead of adapting to the target's new position. We refer to this failure mode as \emph{target-response collapse}.

To mitigate target-response collapse, our key idea is to \textit{decouple the context used to generate a plan from the state in which that plan is executed}. Rather than asking the policy to recover directly from an unfamiliar robot--target configuration, we generate a plan from a familiar canonical robot context, placing the target at the position where the interaction is expected to occur. The resulting canonical plan is then connected to the robot's actual execution state.

The remaining challenge is to determine this future interaction position efficiently. Regenerating a canonical plan whenever the target moves would require repeated costly WAM rollouts. Instead, we predict the interaction in advance: by rolling out future observations and actions beyond its native action chunk, the WAM estimates \emph{when} a key event such as grasp or release would occur before physical execution. Observed target motion then determines \emph{where} the target is expected to be at that time.

We implement this idea through \textbf{Dynamic Predictive Planning (DPP)}, which generates a canonical plan for the predicted interaction and connects it to the robot's ongoing execution. The robot continues acting during planning and subsequently refines the remaining trajectory using fresh observations and its measured state. DPP requires no additional policy training and applies to both joint and cascaded WAMs that can roll out future observations and actions.

Across two WAM backbones, simulation benchmarks, and real-robot tasks, DPP consistently outperforms the evaluated baselines, including methods trained with dynamic data in simulation. Running learned inference on a \textit{single consumer GPU}, DPP makes WAM-based visual prediction practical for real-time planning and execution. These results demonstrate that pretrained manipulation skills can be deployed in dynamic environments without updating the WAM itself. Rather than relearning the skill, DPP adapts its execution to the robot's current state and the target's future interaction point. Our contributions are as follows.
\begin{itemize}[leftmargin=10pt,labelindent=0pt]
\item We identify and empirically characterize \emph{target-response collapse}, a failure mode in which reactive replanning pairs robot states with relocated targets that are unsupported by static demonstrations, causing the policy to favor familiar but incorrect subgoals.

\item We show that decoupling the context used for plan generation from the robot state used for execution allows learned manipulation skills to remain effective under target motion, even when the desired skill is difficult to recover directly from the current observation.

\item We introduce DPP, a backbone-agnostic real-time planning and execution framework that combines interaction-time prediction, target-position prediction, counterfactual canonical planning, and asynchronous plan transitions, enabling dynamic manipulation without additional training on dynamic data.

\end{itemize}

\section{Related Work}
\label{sec:related-work}

\paragraph{Manipulating moving targets.}
Classical dynamic grasping predicts the target's trajectory and plans a reachable grasp under time constraints~\citep{akinola2021dynamic,jia2024dynamic}. Learning-based methods instead add dynamic experience: SIDO augments static demonstrations with counterfactual object poses and actions~\citep{shin2026static}, DynamicManip synthesizes dynamic episodes from one static demonstration~\citep{liao2026dynamicmanip}, and PUMA~\citep{fang2026towards} and DynamicWAM~\citep{lou2026dynamicwam}, our trained baselines, learn motion-aware policies. Others train a module that supplies what one snapshot lacks: AHEAD predicts future visual tokens for a frozen VLA~\citep{syed2026intercepting}, F2F-AP anticipates the observation after system delay~\citep{wei2026f2f}, and TEMPO conditions on motion summaries and proprioceptive history~\citep{feng2026tempo}. DPP enables dynamic manipulation without additional training by changing the context used to invoke a learned skill, mitigating the target-response collapse analyzed in Section~\ref{sec:target-response-collapse}.

\paragraph{Visual prediction and asynchronous execution.}
WAMs fold visual prediction, long used as a planning signal~\citep{finn2017deep,du2024video}, into the action model itself~\citep{LiL-RSS-26,ye2026world}. Fast-WAM and ImageWAM cut deployment cost by omitting explicit future imagination or using image-editing representations~\citep{yuan2026fast,zhang2026imagewam}. DPP instead adapts these backbones to roll out a whole skill, reads off \emph{when} a grasp or release will occur, and extrapolates the target's motion to that moment. Real-Time Chunking continues a committed action prefix while the next chunk is generated~\citep{black2026real}, and AHA-WAM runs a world planner and a feedback policy at different rates~\citep{cai2026aha}. DPP likewise executes while planning, but its activation bridge must also join a trajectory planned from a canonical robot context to the robot's actual state.

\paragraph{Observation editing and skill reuse.}
GenAug and MimicGen broaden training coverage with generated images and demonstrations~\citep{chen2023genaug,mandlekar2023mimicgen}. Closer to DPP are inference-time interventions: ReOI removes distractors before world-model prediction~\citep{chen2025reimagination}, and VLS steers a frozen policy's action sampler with rewards from a vision--language model~\citep{liu2026vls}. DPP edits the target itself, placing it at its predicted interaction position in a familiar robot observation with matching proprioception. This changes only the context used to invoke the skill; execution starts from the live robot state. Its reach is bounded by the backbone's static skill and by how predictably the target moves. Appendix~\ref{app:extended-related-work} extends these comparisons.

\section{Target-Response Collapse: A Failure Mode}
\label{sec:target-response-collapse}

\paragraph{Motivation: dynamic replanning can leave the demonstrated action support.}
Let $S$ denote the physical robot state and $\bm x$ the target position.
A WAM generates an action chunk conditioned on the context
$C(S,\bm x)$:
\begin{equation}
A:=(\bm a_0,\ldots,\bm a_{H-1})\sim p_\theta(\cdot\mid C(S,\bm x)),
\label{eq:analysis-policy-context}
\end{equation}
where $A$ denotes an $H$-step action chunk, and the context is defined by
\begin{equation}
C(S,\bm x)=\big(O,R, I) ~.  %
\end{equation}
Here, $O:=h(S,\bm x,B)$ denotes the current visual observation, with $B$ representing background and camera conditions; $R:=g(S)$ denotes the proprioceptive state, and $I$ is the task instruction.

For a context $C$, we write $\zeta(C)$ for the trajectory generated by recursively querying the frozen WAM. Separately, let \(\zeta_{\mathrm{data}}^{(i)}\) denote the \(i\)-th recorded manipulation demonstration for a stationary target at \(\bm x^{(i)}\). Grouping its recorded actions into chunks, we write
\begin{equation}
\zeta_{\mathrm{data}}^{(i)}:=\big(
S_0^{(i)}, A_0^{(i)}, S_1^{(i)}, A_1^{(i)}, \ldots,
A_{T_i-1}^{(i)}, S_{T_i}^{(i)}
\big),
\qquad
S_0^{(i)} \approx S_0,
\label{eq:analysis-demonstration-context}
\end{equation}
where \(S_t^{(i)}\) is the recorded robot state at chunk boundary \(t\), and \(A_t^{(i)}\) is the corresponding recorded action chunk. In the static-demonstration setting considered here, demonstrations begin from approximately the same initial robot state \(S_0\) and proceed toward their respective targets \(\bm x^{(i)}\).

This demonstration structure induces a nonuniform joint support over robot states and target locations. Near the common initial state \(S_0\), similar robot configurations are paired with many different targets across demonstrations. As execution progresses, the robot state \(S_t^{(i)}\) becomes increasingly specific to the trajectory toward \(\bm x^{(i)}\).

Meanwhile, the training data may contain the contexts
$(S_t^{(i)},\bm x^{(i)})
$ and $(S_t^{(j)},\bm x^{(j)}),
$
without containing the crossed context
$
(S_t^{(i)},\bm x^{(j)}).
$
However, a moving target creates precisely such a crossed context: the robot reaches \(S_t^{(i)}\) while following the action sequence associated with one target location, while the target has meanwhile moved elsewhere. Replanning from this new observation therefore asks the policy to generate behavior from a robot--target combination that may be far outside its demonstrated action support.

\paragraph{Why later robot context can dominate target relocation.}
We analyze this effect under a memoryless planning interface,
\begin{equation}
p_\theta(A\mid C,\mathcal H_k)=p_\theta(A\mid C),
\label{eq:analysis-memoryless}
\end{equation}
where $\mathcal H_k$ denotes prior execution history. In this setting, previous actions affect future behavior only through the current observation and robot state. Importantly, the robot configuration itself carries information about execution progress and can therefore bias the policy toward the continuation associated with that state.

To measure how strongly the model still responds to target relocation, let
$F(\zeta(C))$ denote the predicted grasp location measured from the generated trajectory. For a displaced target
$\bm x'=\bm x+\lambda\bm u$, with $\lambda>0$ and $\|\bm u\|=1$, we define the directional response gain
\begin{equation}
G(S;\bm x,\bm x')=
\frac{\bm u^\top[F(\zeta(C(S,\bm x')))-F(\zeta(C(S,\bm x)))]}{\lambda}.
\label{eq:analysis-context-gain}
\end{equation}
A gain near one indicates that the predicted grasp moves with the target, whereas a gain near zero indicates little directional adjustment.

To isolate the effect of robot context, define
\begin{equation}
\Delta_{R,k}F(\bm x)=F(\zeta(C(S_k,\bm x)))-F(\zeta(C(S_0,\bm x))).
\label{eq:analysis-robot-context-effect}
\end{equation}
The change in target responsiveness between the later and initial robot contexts is
\begin{equation}
G(S_k;\bm x,\bm x')-G(S_0;\bm x,\bm x')
=\frac{\bm u^\top[\Delta_{R,k}F(\bm x')-\Delta_{R,k}F(\bm x)]}{\lambda}
.
\label{eq:analysis-context-response-change}
\end{equation}
A target-independent shift in predicted grasp location cancels in this difference. A reduction therefore reflects a genuine loss of responsiveness to target relocation rather than a uniform change in grasp position.

\begin{wrapfigure}[14]{r}{0.40\textwidth}
    \centering
    \includegraphics[width=\linewidth]{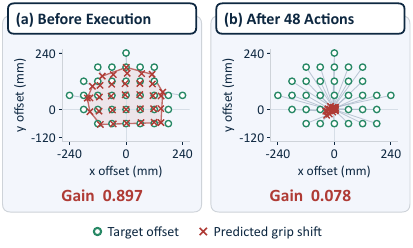}
    \caption{\textbf{Target-response collapse.} Predicted grip shift : before \& after}
    \label{fig:target-response-compact}
\end{wrapfigure}
Figure~\ref{fig:target-response-compact} summarizes the decline in target responsiveness along a model-generated reference trajectory, quantified by the matched target-relocation experiments in Appendix~\ref{app:matched-relocation-probe} (Figure~\ref{fig:prefix-target-response-comparison}). On the same 34-cell static-support set, the predicted grip-response area falls to 1.47\% of its initial value after 48 executed actions. On the same roller-grasping task, ImageWAM also exhibits a decrease in mean target-response gain from 0.885 to 0.063 after 48 executed actions. This trend extends to dual-bottle manipulation: after 24 executed actions, before either gripper contacts an object, relocation success decreases from 100\% for both models to 66.7\% for FastWAM and 72.9\% for ImageWAM, while unshifted controls remain successful at every tested prefix.

\section{Method}
\label{sec:method}

We aim to generate a plan in a context where the frozen WAM can express the required skill, rather than demanding recovery directly from the current robot--target combination.
Toward this goal, Dynamic Predictive Planning (DPP) separates the context $C$ used to generate a plan $\zeta(C)$ from the robot's actual execution state $S_k$. Motivated by Section~\ref{sec:target-response-collapse}, canonical planning uses a familiar robot context with the target represented at its predicted interaction position. The resulting canonical plan $\zeta^{\mathrm{can}}$ is then connected to the robot's ongoing motion.

\begin{figure}[!t]
    \centering
    \includegraphics[width=\linewidth]{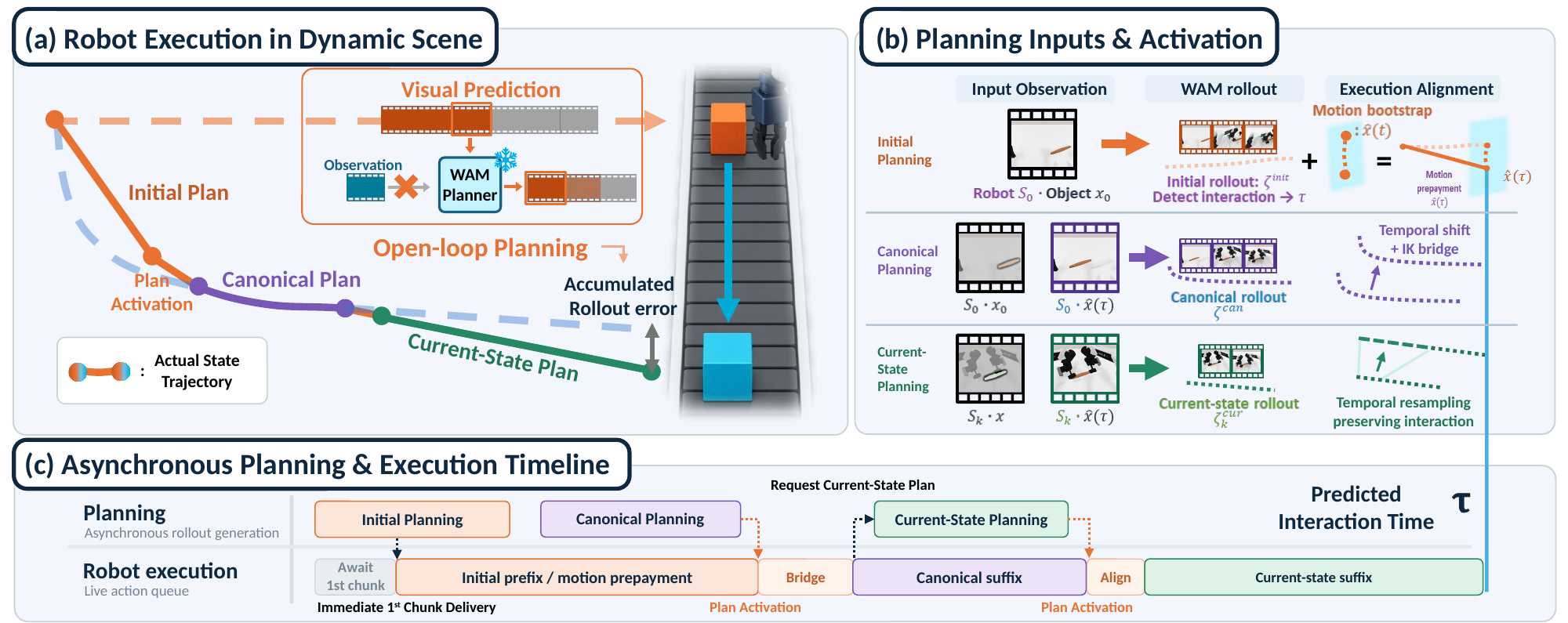}
    \caption{\textbf{DPP pipeline.} An initial rollout predicts interaction timing while available actions begin execution. The canonical planning stage uses the familiar robot context and predicted target position. An activation bridge connects its trajectory to ongoing execution, after which current-state planning uses fresh observations and current robot-state input while retaining the predicted interaction position as its planning target.}
    \label{fig:dpp-pipeline}
    \vspace{-7pt} % Retain one line of separation below the figure.
\end{figure}

Specifically, DPP organizes planning into three stages (Figure~\ref{fig:dpp-pipeline}(a)(b)). The initial planning stage produces a plan $\zeta^{\mathrm{init}}$ to estimate interaction timing and provide actions that can begin executing. The canonical planning stage produces $\zeta^{\mathrm{can}}$ as the reference trajectory for the predicted interaction. The current-state planning stage then produces $\zeta_k^{\mathrm{cur}}$ from fresh observations and current robot-state input to refine the remaining trajectory. In the following, we describe each step in more detail.

\noindent\textbf{Step1. Predict the interaction and prepay execution.}
For the first subtask, initial planning generates $\zeta^{\mathrm{init}}=\zeta(C(S_0,\bm x_0))$ from the observed scene. From this plan, we estimate interaction timing by identifying the first predicted action that satisfies the interaction criterion. In parallel, motion bootstrap uses SAM~2 to segment the target and FoundationPose to track its position across timestamped observations~\citep{ravi2025sam,wen2024foundationpose}. These position estimates are used to fit the target-motion hypothesis $\widehat{\bm x}(\cdot)$, which the base method holds fixed after bootstrap. The interaction timing inferred from the rollout is mapped to the execution clock as
\begin{equation}
    \tau=t_{\mathrm{exec}}+\Delta t\,
    \min\bigl\{j:\phi_q(\bm a_j)=1\bigr\}.
    \label{eq:main-interaction-hypothesis}
\end{equation}
Here, $j$ indexes individual actions in $\zeta^{\mathrm{init}}$, $\Delta t$ is the control time interval, and $t_{\mathrm{exec}}$ is the timestamp of its first executed action. The task-specific predicate $\phi_q$ identifies the interaction event in the predicted action sequence. DPP adopts the event timing inferred from this static-scene rollout as a skill-derived timing prior for dynamic planning. This provides a practical default grounded in the policy's learned behavior, rather than an optimized interaction time for the moving target; alternative timing priors may yield substantially higher success. Mapping this prior to the execution clock gives $\tau$, and evaluating the target-motion hypothesis at $\tau$ gives $\widehat{\bm x}(\tau)$, the target position used for canonical planning.

Once the initial rollout produces its first action chunk, the robot can begin executing that chunk while the WAM continues generating the remaining rollout. The motion prepayment step begins after bootstrap is complete, using the estimated target motion to adjust subsequent actions while the remaining prediction and canonical planning proceed. This allows the robot to approach the target while planning continues.

\noindent\textbf{Step 2. Generate and activate the canonical plan.}
For counterfactual planning, DPP represents the target at a hypothesized position in the WAM's input context. Given an observation $O_k$ and proprioceptive input $R_k$ from robot state $S_k$, we define $\widetilde C(S_k,\bm x')$, a synthesized approximation to the scene context $C(S_k,\bm x')$:
\begin{equation}
    \widetilde C(S_k,\bm x')
    :=\bigl(\operatorname{Edit}(O_k,\bm x'),R_k,I\bigr).
    \label{eq:main-counterfactual-context}
\end{equation}
by applying the visual editing operation $\operatorname{Edit}$   toward the target at $\bm x'$ in RGB--D geometry while preserving the selected robot configuration and scene context. %
Editing is applied consistently across camera views with robot occlusion preserved (Appendix~\ref{app:counterfactual-observation}).

The canonical planning stage uses this construction to combine a familiar robot context with the predicted interaction position. For the first subtask, the familiar reference is the initial robot state $S_0$. Once $\widehat{\bm x}(\tau)$ is available from Step 1, DPP generates the canonical trajectory as
\begin{equation}
    \zeta^{\mathrm{can}}=\zeta\!\left(
        \widetilde C\bigl(S_0,\widehat{\bm x}(\tau)\bigr)
    \right).
    \label{eq:main-counterfactual-plan}
\end{equation}
Since the robot continues executing available actions and reaches a later state $S_k$ during this canonical trajectory generation, the robot must therefore transition from its current state $S_k$ onto the canonical trajectory before following it.
Specifically, DPP selects an entry point in $\zeta^{\mathrm{can}}$ based on the remaining time until $\tau$, without temporally resampling the canonical trajectory. It then constructs an \emph{activation bridge} from the current robot state to the selected point, using inverse kinematics subject to joint-velocity limits (Figure~\ref{fig:dpp-pipeline}(c)). If this connection is accepted, DPP replaces the remaining action queue with the bridge followed by the canonical trajectory from the selected entry point onward. Existing actions continue while this replacement is prepared.

\noindent\textbf{Step 3. Replan from the realized state.}
After the canonical activation bridge has completed, the robot reaches a state $S_k$ close to the canonical trajectory for the predicted interaction position. This is intended to bring the robot--target context closer to the skill's learned support. DPP then uses fresh observations and current robot-state input at $S_k$ to ground the new rollout in the realized state, helping reduce accumulated prediction drift. The candidate continuation is
\begin{equation}
    \zeta_k^{\mathrm{cur}}=\zeta\!\left(
        \widetilde C\bigl(S_k,\widehat{\bm x}(\tau)\bigr)
    \right).
    \label{eq:main-current-plan}
\end{equation}
The plans in Eqs.~\eqref{eq:main-counterfactual-plan} and~\eqref{eq:main-current-plan} use the same rollout procedure but different robot contexts: $S_0$ for canonical planning and $S_k$ for current-state planning. In both planning inputs, the target is represented at $\widehat{\bm x}(\tau)$. This updates the robot context to the realized state while retaining the predicted interaction position. As Section~\ref{sec:target-response-collapse} shows, fresh observations alone do not guarantee responsiveness to target relocation; here, current-state planning refines the remaining trajectory after canonical activation.

For an accepted current-state plan, DPP temporally resamples the joint-position trajectory before the interaction segment to align execution with the remaining time until $\tau$. The interaction segment, including the final approach and grasp or release, is preserved. Base DPP retains the bootstrap motion hypothesis and applies no additional spatial correction to current-state output actions. Optional nonlinear-motion extensions can update motion estimates and spatially correct output actions, as described in Appendix~\ref{app:motion-update-options}.

\noindent\textbf{Continue across subtasks.}
For a task with multiple subtasks, DPP connects plans according to each stage's interaction structure. For stages linked by continuation, the current-state rollout $\zeta_k^{\mathrm{cur}}$ extends beyond the current interaction to predict the next event, and its continuation is reused as the next stage's initial plan. At boundaries designated to start a new planning cycle, DPP generates a new initial plan from a fresh reference observation. Already initialized targets retain their motion hypotheses, while targets that begin moving later are initialized separately. Each stage's event is detected from predicted actions using its corresponding predicate, and its timing is mapped to the execution clock. At a stage transition, a fresh observation and matching proprioception define the next stage's canonical robot reference, replacing $S_0$ in Eq.~\eqref{eq:main-counterfactual-plan}.
Detailed rollout construction, event timing, acceptance, and activation rules are given in Appendix~\ref{app:dpp-implementation}.

\section{Experiments}
\label{sec:evaluation}

We evaluate DPP on dynamic Pick and Pick-and-Place across matched target
positions, velocities, directions, and object configurations. We compare
standard WAM execution, inference-time adaptation, and methods with additional training
under shared scene definitions and success criteria.

\begin{table*}[t]
    \centering
    \small
    \caption{\textbf{RoboTwin2.0-Dyn results.} Each method uses 300 T1 and 192 T2 trials. Success follows the official task predicate; Contact is a separate post hoc diagnostic and may be lower than Success (Appendix~\ref{app:main-results-details}).}
    \label{tab:main-dynamic-results}
    \resizebox{\textwidth}{!}{%
    \begin{tabular}{@{}llcrrrrrr@{}}
        \toprule
        \multirow[c]{2}{*}[-0.6ex]{\shortstack[l]{Backbone /\\Model}}
        & \multirow[c]{2}{*}[-0.6ex]{\shortstack[l]{Inference\\Method}}
        & \multirow[c]{2}{*}[-0.6ex]{\shortstack{Additional\\Dyn. Training}}
        & \multicolumn{2}{c}{Success $\uparrow$}
        & \multicolumn{2}{c}{Contact $\uparrow$}
        & \multicolumn{2}{c}{RC $\uparrow$} \\
        \cmidrule(lr){4-5}\cmidrule(lr){6-7}\cmidrule(l){8-9}
        & & & T1 & T2 & T1 & T2 & T1 & T2 \\
        \midrule
        FastWAM & Refresh-24 (Default) & X & 10.00 & 33.33 & 8.67 & 80.73 & 56.52 & 71.68 \\
                & Refresh-8 & X & 13.00 & 38.54 & 12.67 & 78.65 & 68.97 & 68.88 \\
                & Future-Conditioned (DPP bootstrap) & X & 18.33 & 34.38 & 21.33 & 78.13 & 69.32 & 63.36 \\
                & Future-Conditioned (Oracle) & X & 27.00 & 48.96 & 27.33 & 84.90 & 77.70 & 72.99 \\
        ImageWAM & Refresh-16 (Default) & X & 11.33 & 39.58 & 11.33 & 76.56 & 61.97 & 70.62 \\
                 & Refresh-8 & X & 12.67 & 46.88 & 12.67 & 79.17 & 61.97 & 69.71 \\
                 & Future-Conditioned (DPP bootstrap) & X & 18.33 & 48.96 & 20.33 & 90.63 & 68.14 & 72.20 \\
                 & Future-Conditioned (Oracle) & X & 22.67 & 60.42 & 32.00 & 92.71 & 72.49 & 78.17 \\
        \midrule
        $\pi_{0.5}$ & Default & X & 9.00 & 13.02 & 8.33 & 70.83 & 54.71 & 53.77 \\
        AHA-WAM & Default & X & 7.33 & 14.58 & 5.33 & 83.85 & 53.24 & 58.12 \\
        PUMA & Default & O & 50.33 & 9.38 & 31.67 & 83.85 & 80.54 & 49.51 \\
        DynamicWAM & Default & O & 52.00 & 53.13 & 64.67 & \underline{95.31} & 78.51 & 75.89 \\
        \midrule
        \rowcolor{cyan!7}
        \textbf{FastWAM + VP} & \textbf{DPP (Ours)} & X & \textbf{83.33} & \underline{80.73} & \textbf{94.67} & \textbf{96.35} & \textbf{94.21} & \underline{92.14} \\
        \rowcolor{cyan!7}
        \textbf{ImageWAM + VP} & \textbf{DPP (Ours)} & X & \underline{78.67} & \textbf{91.15} & \underline{93.33} & \textbf{96.35} & \underline{93.43} & \textbf{95.99} \\
        \bottomrule
    \end{tabular}%
    }
    \vspace{-8pt} % Retain one line of separation below the table.
\end{table*}

\subsection{Experimental Setup}

\noindent\textbf{Tasks and protocol.}
We introduce RoboTwin2.0-Dyn, a dynamic manipulation benchmark built on RoboTwin~2.0, and evaluate DPP and the baselines on dynamic variants of four tasks, comprising 300 Bimanual Grasp and 192 Pick-and-Place episodes. We remeasure static support because RoboTwin's predefined evaluation regions need not match the manipulation capabilities of each pretrained backbone. Evaluation scenes span target velocities, directions, and motion configurations within empirically validated static-support regions. Methods share predeclared scenes and simulator seeds, subject to the documented retry exceptions. Static controls at the corresponding final target positions distinguish stationary-task reachability from dynamic execution error. Scene construction, static controls, and retry exceptions are detailed in Appendix~\ref{app:evaluation-protocol}.

Our real-robot evaluation uses an I2RT-based platform with FastWAM trained only on static demonstrations. Task A, Pick (Move) \& Place, requires grasping a moving piggy bank and placing it into a basin. Task B, Pick \& Place (Move), requires placing a goalpost to intercept a moving ball; its static demonstrations place the goalpost at a ball-blocking location. Each method is evaluated over 30 trials per task.

\noindent\textbf{Models and baselines.}
DPP uses frozen FastWAM and ImageWAM backbones with the same motion bootstrap, counterfactual observation synthesis, plan activation, and current-state planning controller; only the planner interface differs. We compare standard and more frequent observation refresh, same-backbone Future-Conditioned inference, $\pi_{0.5}$, AHA-WAM, and additional-training methods PUMA and DynamicWAM. Both Future-Conditioned variants refresh every 8 action steps with a fixed 16-step look-ahead. The Oracle variant uses simulator-rendered future observations. The DPP bootstrap variant estimates motion from observations at steps 0, 2, 4, 6, and 8 before taking policy actions, then combines the estimate with current RGB-D observations to synthesize future inputs without ground-truth rendering or velocity fallback. Neither variant uses DPP's skill-dependent interaction timing or full controller. This comparison tests whether fixed-horizon future observations alone suffice, or whether dynamic manipulation benefits from DPP's interaction-aware planning and execution alignment. The FastWAM/ImageWAM refresh baselines assume zero inference latency. DPP and the same-backbone baselines require no additional policy training. Detailed configurations are provided in Appendix~\ref{app:main-results-details}.

On the real robot, Reactive replans every 10 steps, while Future-Conditioned uses DPP's motion predictor and counterfactual observation synthesis with a fixed 16-step horizon instead of DPP's inferred interaction time. In Task B, DPP and Reactive (Hybrid) share the same policy trained on anticipation-oriented static demonstrations.

\noindent\textbf{Metrics and implementation.}
We report official RoboTwin Task Success, a separate physical-contact diagnostic (Contact), and Route Completion (RC; 0--100). Pick-and-Place success requires completion of both subtasks. All learned DPP inference runs on a single RTX~5090 in simulation and a single RTX~4090 on the real robot; simulation rendering uses a separate GPU. Motion bootstrap and visual planning run asynchronously with nominal 10 Hz control. Metric definitions and runtime details are provided in Appendices~\ref{app:main-results-details} and~\ref{app:runtime-fallback}.

\subsection{Results}

\noindent\textbf{Dynamic Simulation Results.}
Table~\ref{tab:main-dynamic-results} compares task success across both dynamic
settings. In Bimanual Grasp, DPP reaches \textbf{83.33\%} with FastWAM and
\textbf{78.67\%} with ImageWAM. The corresponding Future-Conditioned
DPP bootstrap variants achieve 18.33\% with each backbone, while the Oracle
variants achieve 27.00\% and 22.67\%, respectively.
In Pick-and-Place, DPP reaches \textbf{80.73\%} with FastWAM and
\textbf{91.15\%} with ImageWAM, compared with 34.38\% and 48.96\% for
Future-Conditioned DPP bootstrap, and 48.96\% and 60.42\% for
Future-Conditioned Oracle. DPP achieves the highest observed success rates
across the evaluated methods in both tasks.
We also report success across target-velocity bins. DPP achieves higher success rates
than the evaluated baselines across all target-velocity bins, including the
highest-speed bin.
The full Pick velocity breakdown is reported in
Appendix~\ref{app:velocitywise-results}.

\begin{figure}[!t]
    \centering
    \includegraphics[width=\linewidth]{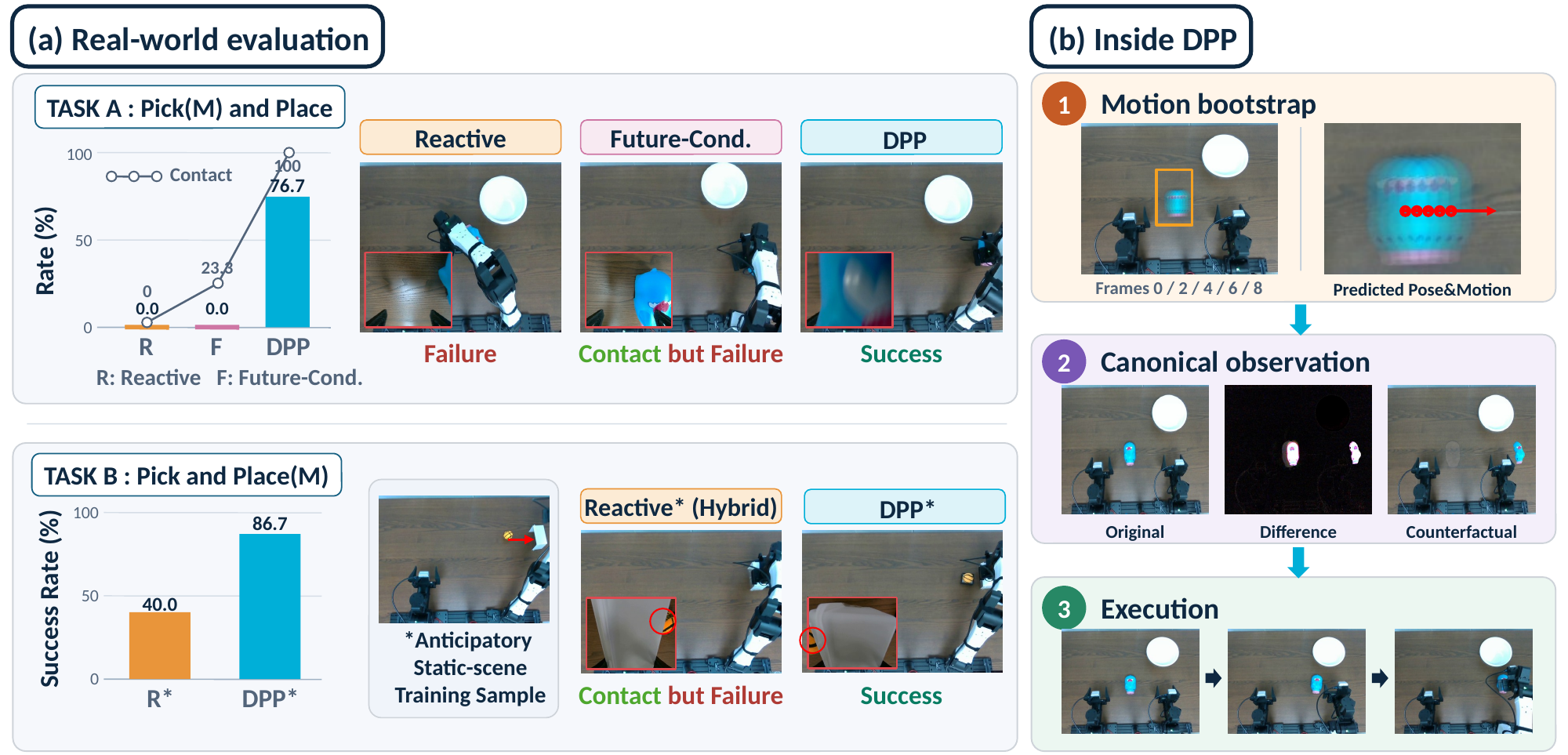}
    \caption{\textbf{Real-world results and DPP execution.} Each method uses 30 trials per task. Bars: success; Task A line: human-evaluated gripper--object contact. Reactive replans every 10 steps; Future-Cond. uses DPP counterfactual observations at $+16$ steps. Task B's starred methods share a policy trained on anticipation-oriented static demonstrations; Hybrid uses reactive inference.}
    \label{fig:real-robot-results}
    \vspace{0pt} % Retain one line of separation below the figure.
\end{figure}

\noindent\textbf{Real-Robot Evaluation.}
Figure~\ref{fig:real-robot-results} summarizes the real-robot results. DPP achieves 76.7\% success in Task A, compared with 0\% for both Reactive and Future-Conditioned, and 86.7\% (26/30) in Task B, compared with 40.0\% for Reactive (Hybrid). Task B illustrates how DPP transfers interaction geometry learned from static demonstrations to the predicted future position of a moving target.

\begin{figure}[!t]
    \centering
    \includegraphics[width=\linewidth]{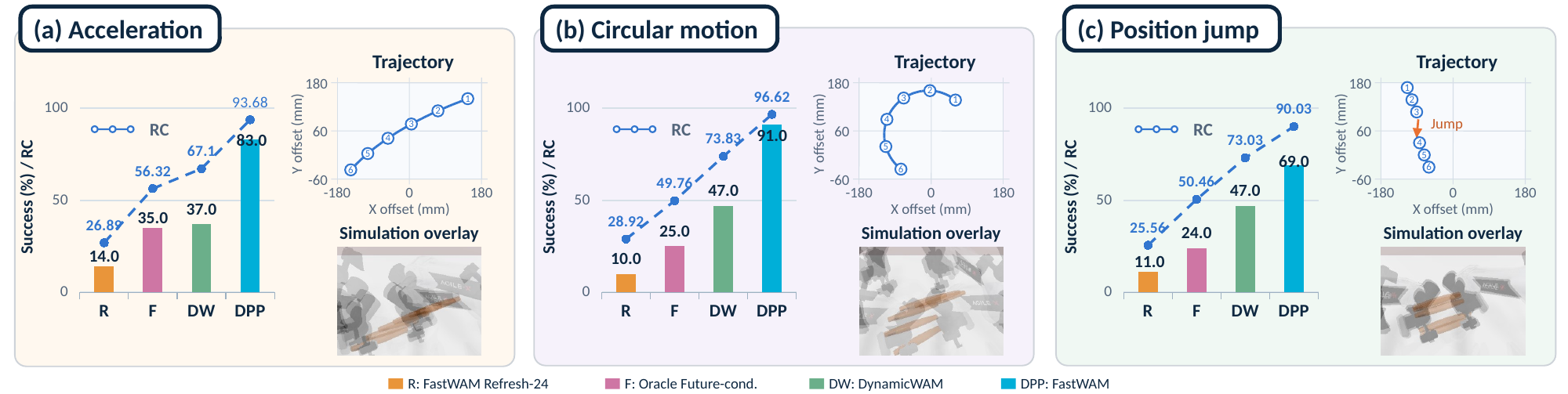}
    \caption{\textbf{Complex target motion.} Each method uses 100 trials per motion family. R: FastWAM Refresh-24; F: $+16$-step oracle observations; DPP: FastWAM with the nonlinear extension and a declared motion-family prior.}
    \label{fig:complex-target-motion}
    \vspace{-10pt} % Retain one line of separation below the figure.
\end{figure}

\subsection{Adaptation to Complex Target Motion.}
\label{sec:complex-target-motion}

Figure~\ref{fig:complex-target-motion} demonstrates DPP's applicability beyond
constant-velocity linear motion, including diverse motion patterns and
trajectory changes modeled as position jumps to represent external
intervention by another robot or a human. For these experiments, we enable
late canonical planning to defer commitment to the motion hypothesis
while accounting for policy latency, improving prediction accuracy, together
with spatial warping of the current-state plan. These extensions
broaden the applicability of prediction-based planning to shared workspaces
with humans and other robots, where object trajectories may change during
handovers, coordinated manipulation, or conveyor-based industrial operations.

\subsection{Ablation Studies}

\noindent\textbf{Planning components.}
Table~\ref{tab:main-component-ablation} reports FastWAM success (\%), Contact (\%), and RC
on the same 300 T1 scenes; RC includes failures. Changes are relative to full DPP,
with $\Delta$ SR and $\Delta$ Contact measured in percentage points. Without initial-plan execution,
the robot waits for canonical planning to complete and executes the resulting
trajectory from the first action;
target motion and the episode budget remain unchanged. Without canonical-state
conditioning, we use the current robot context and its activation
procedure at the same request point, retaining initial execution, the interaction time inferred by initial
planning and its predicted target, and subsequent current-state planning. Both changes
substantially reduce success, while removing current-state planning has a smaller effect.
Interaction-timing and workspace diagnostics are provided in
Appendix~\ref{app:component-ablation-diagnostics}.

\noindent\textbf{Oracle diagnostics.}
On the same scenes, oracle velocity substitutes ground-truth target velocity;
oracle velocity + observation additionally uses simulator-rendered
counterfactual observations
(Table~\ref{tab:main-oracle-diagnostics}). Their gains indicate remaining
headroom in motion bootstrap and observation synthesis. These diagnostic
conditions are excluded from the main comparison. Contact and ImageWAM results
are reported in Table~\ref{tab:dpp-oracle-ablation} in
Appendix~\ref{app:oracle-analysis}.

\noindent\begin{minipage}{\textwidth}
    \makeatletter\def\@captype{table}\makeatother
    \centering
    \caption{\textbf{FastWAM ablations and Oracle diagnostics on Task 1.}}
    \label{tab:main-component-ablation}
    \label{tab:main-oracle-diagnostics}
    \small
    \setlength{\tabcolsep}{0pt}
    \renewcommand{\arraystretch}{1.08}
    \begin{tabular}{@{}p{0.33\textwidth}p{0.10\textwidth}p{0.10\textwidth}p{0.15\textwidth}p{0.13\textwidth}p{0.09\textwidth}p{0.10\textwidth}@{}}
        \toprule
        \textbf{Variant} & \hfill\textbf{SR (\%)} & \hfill$\Delta$ \textbf{SR} & \hfill\textbf{Contact (\%)} & \hfill$\Delta$ \textbf{Contact} & \hfill\textbf{RC} & \hfill$\Delta$ \textbf{RC} \\
        \midrule
        \rowcolor{cyan!7}
        \textbf{Full DPP} & \hfill\textbf{83.33} & \hfill--- & \hfill\textbf{94.67} & \hfill--- & \hfill\textbf{94.21} & \hfill--- \\
        \addlinespace[4pt]
        \multicolumn{7}{@{}l@{}}{\textit{Component ablations}} \\
        w/o initial execution & \hfill50.33 & \hfill$-33.00$ & \hfill46.67 & \hfill$-48.00$ & \hfill83.96 & \hfill$-10.25$ \\
        w/o canonical-state conditioning & \hfill51.67 & \hfill$-31.67$ & \hfill47.33 & \hfill$-47.33$ & \hfill77.96 & \hfill$-16.25$ \\
        w/o current-state planning & \hfill80.00 & \hfill$-3.33$ & \hfill92.33 & \hfill$-2.33$ & \hfill92.82 & \hfill$-1.39$ \\
        \midrule
        \multicolumn{7}{@{}l@{}}{\textit{Oracle diagnostics}} \\
        Oracle target velocity & \hfill92.33 & \hfill$+9.00$ & \hfill95.33 & \hfill$+0.67$ & \hfill97.22 & \hfill$+3.01$ \\
        Oracle target velocity \& observation & \hfill94.00 & \hfill$+10.67$ & \hfill98.00 & \hfill$+3.33$ & \hfill97.90 & \hfill$+3.69$ \\
        \bottomrule
    \end{tabular}
\end{minipage}

\smallskip

\section{Conclusion}
\label{sec:conclusion}

Dynamic manipulation failures can reflect limited access to learned skills rather than missing capabilities. We identify \emph{target-response collapse}: reduced responsiveness to target relocation from later robot contexts. DPP addresses this by separating the planning context from the execution state, using WAM predictions to construct future conditions that elicit existing skills. Simulation and real-robot evaluations show improved dynamic manipulation without additional training on dynamic data, with real-time inference on a single consumer GPU.

These results suggest that dynamic manipulation need not be viewed solely
as a problem of learning new behaviors, since even an already learned behavior
can be more or less accessible depending on the observation context used to
request it. By helping construct such contexts, WAM prediction can serve not
merely to generate future scenes, but to formulate planning problems that an
action model already knows how to solve. The resulting counterfactual
observation can be interpreted as an \emph{implicit retrieval query} for the
model's learned manipulation capabilities.

\paragraph{Limitations and future work.}
\label{sec:limitations}
DPP relies on target motion remaining predictable over the planning
horizon and on the backbone possessing the required static manipulation
skill. Its interaction timing is a practical prior derived from the
learned skill, rather than an optimized choice for the dynamic setting.
Future work could explore uncertainty-aware motion prediction,
timing-prior calibration, and static demonstrations that broaden the
interaction capabilities available for reuse.

\subsection*{AI use statement}
Generative AI tools were used to assist with coding, literature review, and manuscript editing. Their use was limited to supporting the research and writing process, and all AI-assisted outputs were carefully reviewed and, where necessary, revised by the authors. The authors take full responsibility for the accuracy, validity, and final content of the paper.

\subsection*{Ethics statement}
This work investigates robotic manipulation through simulation and real-robot experiments in a controlled laboratory setting. The experiments focus on interactions between robots and inanimate objects, with the aim of improving manipulation under target motion. We do not identify significant ethical concerns specific to the reported methods or experiments, while recognizing that broader deployment requires consideration of physical safety and the intended application.

\subsection*{Reproducibility statement}
Appendix~\ref{app:target-response-validation} specifies the matched target-response
protocol. Appendices~\ref{app:dpp-implementation}
and~\ref{app:component-validation} describe the perception and
counterfactual-editing pipeline, predictive interaction hypothesis, WAM
full-planning interface, and DPP activation and transport rules.
Appendix~\ref{app:dynamic-evaluation} reports the closed-loop evaluation
settings, including checkpoint scope, action horizon, random seeds, motion
conditions, and success criteria. Runtime behavior and measured action queues are described in Appendices~\ref{app:runtime-fallback}
and~\ref{app:representative-runtime-timelines}.

\bibliography{references}
\bibliographystyle{dpp_references}

\newpage
\appendix

\startcontents[appendix]
\section*{Appendix Contents}
\printcontents[appendix]{}{1}{\setcounter{tocdepth}{2}}
\clearpage

\section{Extended Task Evaluation}
\label{app:extended-task-evaluation}

We extend the evaluation beyond the main-text Bimanual Grasp (T1) and
Pick-and-Place (T2) tasks to cover different interaction events and subtask
structures. The additional tasks span object- and arm-specific interaction
timing, direct contact, partially dynamic manipulation,
and the chaining of multiple subtasks involving multiple moving objects.
The suite contains six task types.

\subsection{Task Suite}
\label{app:extended-task-suite}

Table~\ref{tab:extended-task-suite} summarizes the distinct interaction and
planning requirements of each task type. T1 and T2 evaluate coordinated
bimanual interception and the transition from dynamic grasp to dynamic
placement, respectively; T3--T6 extend these requirements to additional
interaction types and subtask structures.

\begin{table}[!ht]
\centering
\small
\caption{Task suite and evaluation objectives. T1 and T2 are the main-text
tasks; T3 contains two RoboTwin variants.}
\label{tab:extended-task-suite}
\vspace{4pt}
\begin{tabular}{@{}lp{0.18\linewidth}p{0.28\linewidth}p{0.39\linewidth}@{}}
\toprule
ID & Task type & Task / variants & Evaluation objective \\
\midrule
T1 & Bimanual Grasp & \texttt{grab\_roller} & Coordinated two-arm interception of one moving target. \\
T2 & Pick-and-Place & \texttt{move\_pillbottle\_pad}; \texttt{place\_empty\_cup}; \texttt{place\_phone\_stand} & Transitioning from grasp of a moving object to release at a moving placement target. \\
T3 & Bottle Picking & \texttt{pick\_dual\_bottles}; \texttt{pick\_diverse\_bottles} & Object- and arm-specific timing and future target positions. \\
T4 & Bell Clicking & \texttt{click\_bell} & Contact-based interaction-time estimation. \\
T5 & Block Handover & \texttt{handover\_block} & Integrating a dynamic stage into a multi-subtask sequence. \\
T6 & Block Stacking & \texttt{stack\_blocks\_two} & Chaining multiple subtasks involving multiple moving objects. \\
\bottomrule
\end{tabular}
\end{table}

\paragraph{Coordinated bimanual interception: T1.}
Bimanual Grasp requires both arms to engage the same moving object. Planning
must coordinate the two end-effectors around a shared interaction time and
future target position while preserving the bimanual grasp geometry. This
evaluates whether a learned two-arm manipulation skill can be connected to a
moving target as a coordinated interaction.

\paragraph{Dynamic grasp-to-placement transition: T2.}
Pick-and-Place requires grasping a moving object and subsequently releasing
it at a moving placement target. The relevant target and interaction
predicate change between stages, from gripper closing for grasp to opening
for release. Planning must therefore transfer the state reached after grasp
into a new reference and predict the placement target's position at the
release time. Co-motion and counter-motion scenes evaluate this transition
under different relative motions of the object and placement target.

\paragraph{Object- and arm-specific timing: T3.}
Bottle Picking assigns separate objects to the two arms. Rather than assuming
simultaneous interaction, planning distinguishes each object's future target
position and each arm's interaction time. Its two variants evaluate this
structure under different object configurations.

\paragraph{Contact-based interaction: T4.}
The main-text interaction-time definition uses the first predicted action
satisfying the subtask predicate $\phi_q$. This formulation is not restricted
to gripper closure: the predicate can also be evaluated on robot poses derived
from predicted actions. T4 uses a predicate corresponding to bell contact.
This task extends the same interaction-time formulation to direct contact.

\paragraph{Partially dynamic subtask sequences: T5.}
Block Handover begins by grasping a moving block and then transfers it to the
other arm. Dynamic-target manipulation is thus one part of a longer task,
rather than a property required of every subtask. The evaluation examines
whether the state reached after the dynamic grasp can be connected to the
subsequent handover and completion of the overall sequence.
The handover evaluation uses the predicted grasp boundary to initiate native
current-observation replanning, retaining queued actions until its result arrives.

\paragraph{Multiple dynamic objects and subtasks: T6.}
Block Stacking requires a sequence of subtasks involving multiple moving
objects. Each stage identifies its target and interaction event, and the
state reached after the preceding subtask supplies the next planning
reference. This evaluates continuous planning and execution across several
objects and interactions.

\subsection{Task-specific Interaction-time Detection}
\label{app:extended-task-events}

Interaction times follow Eq.~\ref{eq:main-interaction-hypothesis}: the first
predicted action satisfying the task-specific predicate $\phi_q$ is mapped to
the execution clock. These times determine the target-position hypotheses
and, for sequential planning stages, the timing of the planning transition.
Table~\ref{tab:extended-task-events} specifies how each interaction is detected
from predicted actions. The detectors do not use simulator contact signals
and do not themselves certify physical contact or task success.

Grasp and release are detected from predicted gripper commands. Contact-related
interaction times are detected from TCP poses obtained by forward kinematics
of the predicted joint actions. These planning predicates are distinct from
the simulator's physical handoff and success rules.

\begin{table}[!ht]
\centering
\small
\caption{Task-specific interaction-time detection from predicted actions.
The table describes planning predicates, not simulator success conditions.}
\label{tab:extended-task-events}
\begin{tabular}{@{}lp{0.24\linewidth}p{0.65\linewidth}@{}}
\toprule
Task & Subtask / interaction & Detection condition in predicted actions \\
\midrule
T1 & Bimanual grasp & First bilateral gripper-closing event. \\
T2 & Grasp & First sustained closing event of the active-arm gripper. \\
 & Release & First sustained gripper-opening event after the predicted grasp and full closure. \\
T3 & Each bottle grasp & First sustained gripper-closing event, detected independently for each assigned arm and object. \\
T4 & Bell press & First action whose predicted TCP height satisfies the press condition relative to the calibrated tabletop. \\
T5 & Initial block grasp & First sustained closing event of the giving-arm gripper. Subsequent handover uses native current-observation replanning toward fixed destinations. \\
T6 & Each block grasp & First sustained closing event of the assigned-arm gripper. \\
 & Each block release & First upward crossing of the gripper-opening boundary after the predicted grasp. \\
\bottomrule
\end{tabular}
\end{table}

\subsection{Evaluation Results}
\label{app:extended-task-results}

The comparator cohorts use the supplied task-specific target-motion sets and
episode budgets. Success requires
completion of the full task, including all required subtasks; detecting an
individual interaction event does not count as task success. We report
Success (\%), following the main evaluation.

Table~\ref{tab:extended-task-results} groups tasks by column and methods by row. T3 pools successes across the Dual and Diverse Bottle Picking variants
(100 trials each), yielding one success rate over 200 trials. Comparator T1--T2 entries reproduce Table~\ref{tab:main-dynamic-results}.
FastWAM uses Refresh-24; FC-FastWAM uses fixed $+16$-step oracle observations.
DPP uses the frozen FastWAM backbone in all columns shown here, with 300
trials for T1, 192 for T2, 200 for T3, and 100 per remaining additional task.

\begin{table}[!ht]
\centering
\small
\setlength{\tabcolsep}{0pt}
\renewcommand{\arraystretch}{1.12}
\setlength{\belowcaptionskip}{5pt}

\caption{Success (\%) across the task suite. Higher is better.
Trials per method are 300 for T1, 192 for T2, 200 for T3
(100 per Bottle Picking variant), and 100 per remaining additional task.
-- denotes omitted PUMA/DynamicWAM results on tasks absent from DOMINO.
DPP uses the FastWAM backbone.}
\label{tab:extended-task-results}

\begin{tabular}{@{}
    >{\raggedright\arraybackslash}p{0.28\linewidth}
    >{\centering\arraybackslash}p{0.11\linewidth}
    >{\centering\arraybackslash}p{0.13\linewidth}
    >{\centering\arraybackslash}p{0.15\linewidth}
    >{\centering\arraybackslash}p{0.09\linewidth}
    >{\centering\arraybackslash}p{0.13\linewidth}
    >{\centering\arraybackslash}p{0.11\linewidth}@{}}
\toprule
Method & T1 & T2 & T3 & T4 & T5 & T6 \\
 & Grasp & Pick--Place & Bottle Picking & Bell & Handover & Stacking \\
\midrule
FastWAM (Refresh-24) & 10.00 & 33.33 & 1.50 & 3.00 & 11.00 & 0.00 \\
FC-FastWAM (Oracle) & 27.00 & 48.96 & 24.00 & 11.00 & 58.00 & 8.00 \\
DynamicWAM & 52.00 & 53.13 & -- & 5.00 & 24.00 & -- \\
PUMA & 50.33 & 9.38 & -- & 4.00 & 19.00 & -- \\
$\pi_{0.5}$ & 9.00 & 13.02 & 3.50 & 0.00 & 12.00 & 0.00 \\
\midrule
\rowcolor{cyan!7}[0pt][0pt]
DPP (FastWAM) & 83.33 & 80.73 & 72.00 & 63.00 & 91.00 & 64.00 \\
\bottomrule
\end{tabular}
\end{table}

\section{Target-Response Validation Protocol}
\label{app:target-response-validation}
\label{app:matched-relocation-probe}

\subsection{Matched Prefix Target-Relocation Probe}

\begin{figure}[t]
\centering
\includegraphics[width=\linewidth]{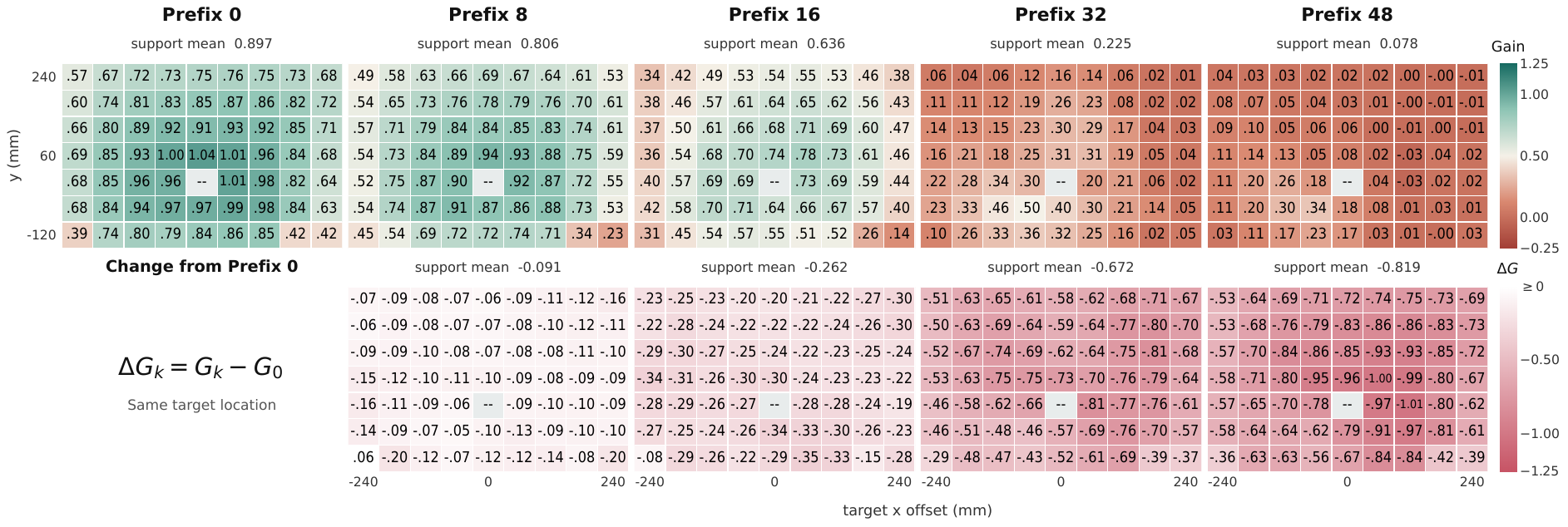}
\caption{\textbf{Target-response collapse under matched target relocation.}
Directional grip-response gain (top) and its change from prefix 0 (bottom).
Maps show all 63 target cells; reported means use the same 33 noncentral
cells in the fixed static-support mask. Gray denotes undefined gains.
Color scales differ by row.
}
\label{fig:prefix-target-response-comparison}
\label{fig:a40-prefix-grip-response-gain}
\label{fig:target-pair-directional-gain-grid}
\end{figure}

We evaluate target responsiveness on the \texttt{grab\_roller}
task using a shared $9\times7$ target grid spanning
$x\in[-240,240]$\,mm and $y\in[-120,240]$\,mm at 60\,mm spacing.
Three static closed-loop executions per cell define an empirical support mask: a cell is
in static support only when all three executions succeed. This gives 34 robust
cells out of 63. The mask is fixed before the matched-prefix experiment and is
used only to stratify outcomes; it is not exposed to the policy.

For each prefix $k\in\{0,8,16,32,48\}$, measured in executed actions, we restore the matched robot state,
relocate the target to every grid cell, and query a new plan from the same
frozen FastWAM checkpoint. Each condition has three paired-repeat queries and
physical simulator rollouts, giving $5\times63\times3=945$ trials. Matching the
checkpoint, robot prefix, target grid, and repeat count isolates how the
remaining action suffix responds to target relocation as execution advances.

The static calibration uses 189 trials, replanning every 8 actions with a
400-action limit. Policy seeds are 42, 43, and 44; simulator seeds are 100000,
100001, and 100002. Each repeat generates one canonical trajectory, whose
states at the five prefixes are restored before each target relocation.
The instruction and query seed are held fixed within a repeat. Prefix queries
use up to four 32-action autoregressive chunks, stopping after a chunk that
contains the first predicted bilateral closure. The trajectory measurement
$F(\zeta(C))$ is the grip midpoint computed by forward kinematics at that closure,
or at the last predicted action if no closure is detected. Physical evaluation replays the cached predicted actions
and, when needed, continues in 32-action autoregressive chunks using only the
preceding predicted observation and joint state. Each prefix receives the same
400-action budget after relocation and terminates on official task success or
the action limit; predicted closure does not terminate execution. The 945 physical rollouts use the cached gain-query predictions; the
189 static calibration trials separately define the fixed support mask.

\paragraph{Directional grip response gain.}
Using the notation of Section~\ref{sec:target-response-collapse}, let
$\bm x$ be the central control target and
$\bm\delta=\bm x'-\bm x$ a target relocation. For a paired rollout comparison,
the displacement of the predicted bilateral grip midpoint is
\[
\bm r_k(\bm\delta)=F(\zeta(C(S_k,\bm x')))-F(\zeta(C(S_k,\bm x))).
\]
The corresponding directional response uses the gain defined in the main text:
\begin{equation}
    G_k(\bm\delta)
    =
    \frac{\bm r_k(\bm\delta)^\top\bm\delta}
         {\lVert\bm\delta\rVert_2^2}
    =G(S_k;\bm x,\bm x').
    \label{eq:a40-grip-response-gain}
\end{equation}
Reported response vectors and gains are averaged over the three paired repeats
for each condition. Direction and response-hull summaries use the mean response vectors.
A gain near one means that the predicted grip point follows the target
relocation; a gain near zero means that the target moves but the predicted grip
point remains nearly fixed. Negative gain indicates a response in the wrong
direction. The zero-displacement control has no defined gain and is excluded
from gain summaries.

The matched-cell comparison in Figure~\ref{fig:prefix-target-response-comparison} uses
$\Delta G_k(\bm\delta)=G_k(\bm\delta)-G_0(\bm\delta)$, retaining the same
target relocations at each robot prefix. Its mean is computed over the same
fixed support mask, excluding the undefined central control.

Figure~\ref{fig:prefix-target-response-comparison} shows the directional
gain maps. The fixed support mask contains 34 cells; gain and direction-cosine
means use the 33 noncentral cells. Mean gain decreases from 0.897 at prefix 0
to 0.806, 0.636, 0.225, and 0.078 at prefixes 8, 16, 32, and 48.
The corresponding mean direction cosines are 0.999, 0.998, 0.997, 0.930, and
0.649. Direction cosine is computed from the mean paired response at each cell.
Directional alignment largely persists while response magnitude weakens,
with a larger loss of alignment at prefix 48.

\paragraph{Target grid to predicted grip-point contraction.}
In addition to directional gain, we measure how the spatial extent of
target-wise mean predicted grip points changes with execution prefix. We plot
every target--response pair and measure the convex-hull area of the
repeat-averaged response points over the same static-support cells:
\begin{equation}
    R_k^{\mathrm{area}}
    =
    \frac{
      \operatorname{Area}
      \left(\operatorname{Hull}\{\bm r_k(\bm\delta):
      \bm\delta\in\mathcal S\}\right)
    }{
      \operatorname{Area}
      \left(\operatorname{Hull}\{\bm r_0(\bm\delta):
      \bm\delta\in\mathcal S\}\right)
    },
    \label{eq:a40-response-area-retention}
\end{equation}
where $\mathcal S$ is the fixed 34-cell canonical support set.

\begin{figure*}[!t]
    \centering
    \includegraphics[width=0.99\textwidth]{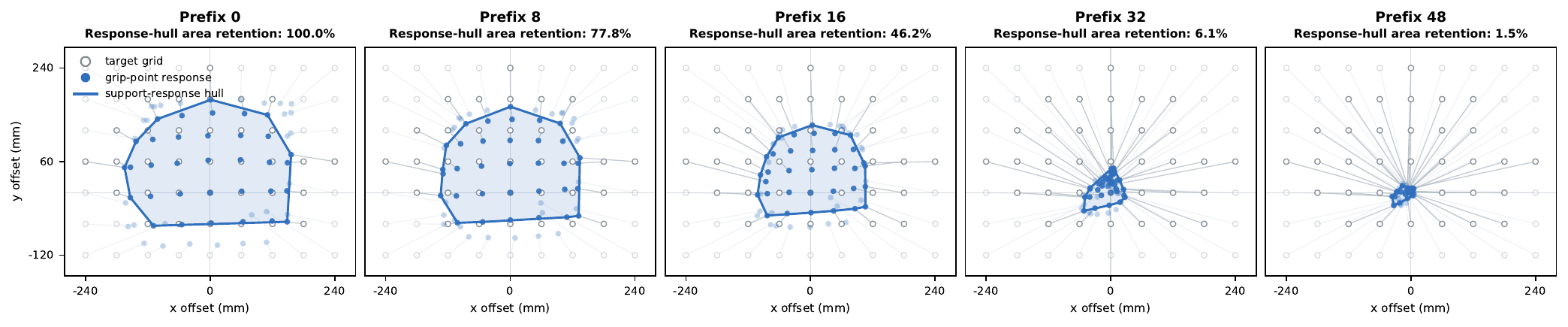}
    \caption{\textbf{Contraction of predicted grip-point responses across execution prefixes.}
    White circles are target offsets, blue circles are control-relative bilateral
    grip responses, and connecting segments preserve each matched pair.
    The blue hull is computed from mean predicted responses over the fixed
    34-cell static-support set. Panel headings report its area relative to
    prefix 0. This measures contraction of predicted responses, rather than
    a change in physically reachable workspace.}
    \label{fig:a40-target-grip-point-contraction}
    \label{fig:matched-target-response-collapse}
\end{figure*}

Figure~\ref{fig:a40-target-grip-point-contraction} shows the target
relocations and mean grip-point responses relative to the paired central
control. Response-area retention falls from 100\% at prefix 0 to 77.82\%,
46.24\%, 6.12\%, and 1.47\% at prefixes 8, 16, 32, and 48.
Mean planar target-to-grip error rises from 23.4 to 34.4, 58.3, 118.0, and
136.1\,mm. Error is the per-trial XY distance between the predicted grip
midpoint and the relocated target, averaged over all 34 support cells and
three repeats. The late-prefix gain reduction is accompanied by contraction
of the spatial extent of target-wise mean predicted responses.

\begin{table*}[!t]
    \centering
    \small
    \caption{Matched-prefix target-response and task-success summary.
    Gain and cosine use 33 noncentral support cells; area and error use all
    34 support cells. SR uses all 63 grid cells, with three repeats per cell.}
    \label{tab:a40-prefix-support-summary}
    \begin{tabular}{rrrrrr}
        \toprule
        Prefix & Mean gain & Direction cosine & Area retention &
        XY error (mm) & SR \\
        \midrule
        0 & 0.897 & 0.999 & 100.00\% & 23.4 & 95.77\% \\
        8 & 0.806 & 0.998 & 77.82\% & 34.4 & 84.13\% \\
        16 & 0.636 & 0.997 & 46.24\% & 58.3 & 67.20\% \\
        32 & 0.225 & 0.930 & 6.12\% & 118.0 & 30.16\% \\
        48 & 0.078 & 0.649 & 1.47\% & 136.1 & 19.58\% \\
        \bottomrule
    \end{tabular}
\end{table*}

\FloatBarrier

\paragraph{Reduced target responsiveness as execution advances.}
Across all 63 target-grid cells and with a 400-action execution budget, physical
grasp success decreases from 95.77\% at prefix 0 to 19.58\% at prefix 48.
Success is evaluated using the simulator's official task-success criterion.
Response metrics on the fixed static-support set and physical success over
the full grid both decrease as execution advances; these summaries use
different evaluation sets. In particular, response magnitude initially
decreases while the response direction remains aligned with the target
relocation. Thus, even when the policy responds in the direction of the new
target, it does not shift the predicted grip point sufficiently; as execution
advances, target-wise mean predicted grip points concentrate within a narrower region.

We define this pattern as \emph{target-response collapse}. Responsiveness
decreases across execution prefixes even for the same target locations that
repeatedly succeed under static conditions. Target-location reachability
alone therefore does not explain the policy's ability to adapt. Target
responsiveness depends on the joint configuration of the current robot state
and target position, and updating the target after execution has progressed
does not by itself guarantee a sufficient action correction. These results
support planning approaches that account for robot state and target position
jointly.

This experiment validates the behavioral limitation; identifying the policy's
internal mechanism that produces it is outside its scope.

\subsection{Additional Models and Tasks}
\label{app:additional-target-response}

\paragraph{Roller grasping with ImageWAM.}
We repeat the matched-prefix probe with ImageWAM on the same $9\times7$
relocation grid and the same three paired scene/policy seeds. Static
closed-loop evaluation, refreshed every eight actions, succeeds in 169/189
trials and identifies 49 cells successful in all three repeats. This
model-specific mask is fixed before the prefix experiment. Mean gain uses
its 48 noncentral cells; success rate uses all 189 trials per prefix.
ImageWAM generates native 16-action chunks, stopping the gain query after
the first chunk containing predicted bilateral closure below 0.2, up to
eight chunks. The grip midpoint is measured at that closure.
The model-generated canonical trajectory
continues to at least 48 actions to supply every prefix state. Physical
evaluation uses the cached prediction and autoregressive continuation from
predicted images and proprioception, terminating on official task success
or 400 post-relocation actions. Table~\ref{tab:additional-target-response}
shows mean gain decreasing from 0.885 to 0.063 and success from 98.41\% to
18.52\% between prefixes 0 and 48.

\paragraph{Dual-bottle grasping with FastWAM and ImageWAM.}
We relocate both bottles by the same planar displacement on a $5\times5$
grid, with $\Delta x\in\{-6,-3,0,3,6\}$\,cm and
$\Delta y\in\{-3,-1.5,0,1.5,3\}$\,cm. The canonical bottle centers are
$(\pm17.2169,6.6690)$\,cm. We use scene seed 8840000 and policy seeds 42
and 43. Static screening uses closed-loop native chunks of 32 actions for
FastWAM and 16 for ImageWAM. All 25 cells succeed in both static trials for
both models, defining a shared support mask. At the reported prefixes
$k\in\{0,8,16,24,32,40\}$, each relocated trial shares its canonical
robot state with the same-model, same-seed, same-prefix no-shift control.
The initial autoregressive prediction spans 128 actions: four 32-action
chunks for FastWAM or eight 16-action chunks for ImageWAM. Continuation
uses predicted images and proprioception, with a 250-action
post-relocation budget and official task-success termination.

For each arm, the response endpoint is its predicted TCP position at the
first gripper closure below 0.9 lasting two consecutive predicted actions.
We apply Equation~\ref{eq:a40-grip-response-gain} to each arm's endpoint
displacement and average over both arms, both seeds, and the 24 noncentral
cells. Success rate likewise excludes the center, using 48 trials per
model and prefix; all no-shift controls succeed. Both additional probes
use ray tracing with OIDN, 32 samples per pixel, and depth 8. Prefixes
count executed policy actions, with native TOPP determining physical duration.

At prefix 24, both arms remain contact-free in both models and seeds,
yet FastWAM and ImageWAM success rates have fallen to 66.67\% and 72.92\%,
with mean gains of 0.511 and 0.712. At prefix 32, ImageWAM has one-arm
contact while FastWAM remains contact-free. Table~\ref{tab:dual-bottle-target-response} reports
results through prefix 40, where success falls to 33.33\% and 41.67\%,
with mean gains of 0.395 and 0.475, respectively. Prefix 48 is omitted
because the bottles have already been grasped, changing the relocation
condition. Reduced target responsiveness therefore appears before contact
in this task; the later-prefix results also reflect the evolving contact state.

\begin{table}[!htbp]
    \centering
    \small
    \caption{\textbf{Roller target response with ImageWAM.}
    SR uses 189 trials per prefix; gain uses 48 noncentral static-support
    cells and three repeats.}
    \label{tab:additional-target-response}
    \begin{tabular}{@{}rrr@{}}
        \toprule
        Prefix & SR (\%) & Gain \\
        \midrule
        0  & 98.41 & 0.885 \\
        8  & 94.18 & 0.817 \\
        16 & 78.31 & 0.722 \\
        32 & 36.51 & 0.274 \\
        48 & 18.52 & 0.063 \\
        \bottomrule
    \end{tabular}
\end{table}

\begin{table}[!htbp]
    \centering
    \small
    \caption{\textbf{Dual-bottle target response through prefix 40.}
    SR uses 48 shifted trials per model and prefix; gain averages both arms
    over 24 noncentral cells and two repeats. Prefix 48 is omitted because
    the bottles have already been grasped.}
    \label{tab:dual-bottle-target-response}
    \begin{tabular}{@{}rrrrr@{}}
        \toprule
        & \multicolumn{2}{c}{FastWAM}
        & \multicolumn{2}{c}{ImageWAM} \\
        \cmidrule(lr){2-3}\cmidrule(l){4-5}
        Prefix & SR (\%) & Gain & SR (\%) & Gain \\
        \midrule
        0  & 100.00 & 1.054 & 100.00 & 1.020 \\
        8  & 100.00 & 0.953 & 100.00 & 0.992 \\
        16 & 100.00 & 0.758 & 100.00 & 0.895 \\
        24 & 66.67 & 0.511 & 72.92 & 0.712 \\
        32 & 47.92 & 0.459 & 50.00 & 0.637 \\
        40 & 33.33 & 0.395 & 41.67 & 0.475 \\
        \bottomrule
    \end{tabular}
\end{table}

\FloatBarrier
\section{Implementation Details}
\label{app:dpp-implementation}
\label{app:planning-execution-details}

\subsection{Overview}

This section describes the common DPP procedures for constant-velocity
targets and their composition across the task suite. Numerical controller
settings refer to the base configuration unless stated otherwise. The nonlinear
motion extension enables additional observation and correction options,
specified in Appendix~\ref{app:motion-update-options}; it does not change
the base configuration.

Dynamic Predictive Planning (DPP) separates the context used to generate a
plan from the physical state from which it is executed. Counterfactual
planning pairs a familiar robot context with a predicted future target
position to invoke the pretrained WAM's existing manipulation capabilities.
Here, canonical specifies the selected robot reference, whereas
counterfactual specifies the hypothesized target position. The initial planning stage
and current-state planning make the resulting canonical plan usable during
ongoing execution, as summarized in Fig.~\ref{fig:dpp-pipeline}.

The initial planning stage and motion bootstrap run concurrently. The initial rollout
provides both a skill-dependent interaction-time estimate and executable
action chunks. When enabled for the active stage, motion prepayment adjusts available
initial actions using the estimated target motion while canonical planning
is prepared. Bootstrap uses
timestamped startup RGB--D observations; object detection compares an
empty-background reference with the reserved initial observation, while later
tracking samples may arrive after execution starts. The timing and motion
estimates select the future target position for the canonical query. After
the activation bridge completes, current-state planning reanchors the
continuation to fresh observations and current robot-state input, reducing
reliance on recursively predicted robot states while retaining the selected
future target. Only an accepted candidate replaces the active suffix.

The episode state machine, action queue, observation synthesis, plan acceptance, and plan activation are shared across backbones. FastWAM returns one native 32-action segment, whereas the ImageWAM adapter combines two 16-action chunks into the same logical 32-action interface. Thus, only planner I/O is backbone-specific.

\paragraph{WAM interface and generated trajectories.}
At each planning step, the frozen WAM takes a visual observation $O$,
proprioceptive input $R=g(S)$, and instruction $I$, and predicts an action
chunk $A$ of length $H$ together with its terminal visual observation:
\begin{equation}
(A,\widehat O)=\mathcal W_\theta(O,R,I).
\label{eq:method-wam-interface}
\end{equation}
We write $C=(O,R,I)$ and use $\zeta(C)$ for a trajectory generated by
recursively querying this model, distinct from the recorded demonstrations
$\zeta_{\mathrm{data}}^{(i)}$ introduced in Section~\ref{sec:target-response-collapse}.
We use \emph{planning} for a generation
stage and \emph{plan} for its output. Initial, canonical, and current-state
planning use the same interface with different contexts.
The index $t$ counts generated chunks, $j$ counts actions within their
concatenation, and $k$ indexes live execution contexts. Physical time $s$
is measured in seconds on the target-motion clock, with control time interval
$\Delta t$. An unaccented $O_k$ is a real observation, $\widehat O_t$ a
model-predicted observation, and $\widetilde O_k$ a counterfactual observation.
The initial observation $O_0$ is reserved before the first action; its
capture time need not coincide with the execution start time.

\subsection{initial planning and motion bootstrap}
\label{app:initial-planning-bootstrap}
\label{sec:initial-planning-bootstrap}

For the first subtask, DPP performs initial planning from the unmodified initial
observation $O_0$, proprioceptive input $R_0$, and instruction $I$, producing
$\zeta^{\mathrm{init}}=\zeta(C(S_0,\bm x_0))$. Starting from
$\widehat O_0=O_0$ and $\widehat R_0=R_0$, the open-loop recurrence is
\begin{equation}
(A_t,\widehat O_{t+1})=
\mathcal W_\theta(\widehat O_t,\widehat R_t,I).
\label{eq:method-initial-rollout}
\end{equation}
The predicted terminal image and final absolute joint-position command in
$A_t$ become the next chunk's visual and proprioceptive inputs, respectively;
no new sensor observation is incorporated into this rollout. The first
segment can execute while further generation and motion bootstrap continue.
DPP applies a
task-specific interaction predicate $\phi_q$ to the predicted action sequence.
Grasp and release predicates use gripper commands, while press
predicates use TCP poses obtained through forward kinematics. Predicates may
also require a preceding event, such as grasp before release. The first qualifying action defines the interaction time:
\begin{equation}
\tau_q = t_{\mathrm{ref},q} + \Delta t\,j_q,
\label{eq:method-interaction-step}
\end{equation}
where $j_q$ is the one-based action count through the detected event and
$t_{\mathrm{ref},q}$ is the time immediately before the rollout's first action.
Thus, an event at zero-based array index $i$ has offset $(i+1)\Delta t$;
the interaction-time offset $j$ in Eq.~\ref{eq:main-interaction-hypothesis}
uses this same convention. For the first initial plan,
this reference is $t_{\mathrm{exec}}$; reused continuations account for the
alignment of the active prefix. Each relevant object or arm retains its own
interaction time and target-position hypothesis. These planning events do
not certify physical contact or task success and do not require simulator
contact or goal-region signals. Appendix~\ref{app:extended-task-events}
summarizes the task-specific predicates. For readability, the following
single-target expressions omit the interaction index.

The motion bootstrap process separates object initialization from motion estimation. Foreground components from the empty-background reference and initial RGB--D observation provide box prompts for SAM~2. Depth conditioning removes table-plane and background support from the selected mask. Its masked RGB--D points form a temporary proxy mesh, which FoundationPose tracks across the buffered startup observations without using a task CAD/GLB, simulator actor mask, or simulator object pose. The tracking sequence is causal but is not restricted to pre-execution frames: initial planning may publish actions before all bootstrap observations have arrived. Only the RGB--D stability gate for declaring static motion is restricted to observations captured before robot execution.

\paragraph{Bootstrap sampling and static-scene gate.}
The base motion-fit buffer requests five observations at offsets
$0,2,4,6,8$ policy steps on the capture clock. These are sampling targets;
the motion fit uses actual timestamps, not an assumed uniform frame rate.
The static-scene test separately requires five pre-execution RGB--D frames.
For each consecutive pair, a pixel is marked changed if its maximum absolute
RGB-channel difference is at least 18 on the 8-bit scale, or its valid-depth
difference is at least $0.004\,\mathrm{m}$. A changed pixel is coherent if at
least three pixels in its $3\times3$ neighborhood, including itself, are
changed. A pair is dynamic when it contains at least 64 coherent changed
pixels. Static motion is declared only when all four pairs pass the stability
test; insufficient frames do not establish a static scene.

Given tracked positions $\widehat{\bm x}_i$ and their actual capture timestamps
$s_i$ in seconds, DPP estimates the bootstrap velocity by
\begin{equation}
(\widehat{\bm b},\widehat{\bm v}_{\mathrm{boot}})
=
\arg\min_{\bm b,\bm v}
\sum_i
\left\|\widehat{\bm x}_i-(\bm b+\bm v s_i)\right\|_2^2.
\end{equation}
Here, $\bm b$ is the position intercept, distinct from an action command.
The fitted $\widehat{\bm v}_{\mathrm{boot}}$ has units of meters per second;
velocity expressed in millimeters per policy step is
$10^3\Delta t\,\widehat{\bm v}_{\mathrm{boot}}$.
For the planar-motion setting used in our experiments, we set
$\hat v_{\mathrm{boot},z}=0$ and retain the bootstrap target height. If both
pre-execution RGB and depth are stable, DPP declares a static scene and sets
$\widehat{\bm v}_{\mathrm{boot}}=\bm 0$.

Write the fitted trajectory as
\begin{equation}
\widehat{\bm x}(s)=f(s;\widehat{\bm\theta}),
\label{eq:method-target-motion}
\end{equation}
where $f$ is the motion family and $\widehat{\bm\theta}$ its fitted parameters.
In the base constant-velocity setting,
$f(s;\widehat{\bm\theta})=\widehat{\bm b}+\widehat{\bm v}_{\mathrm{boot}}s$.
The interaction hypothesis combines $\tau$ with this trajectory
on the same clock to obtain the predicted interaction position,
\begin{equation}
\widehat{\bm x}(\tau)
=\widehat{\bm b}+\widehat{\bm v}_{\mathrm{boot}}\tau.
\end{equation}
Each target's motion hypothesis is frozen after its initialization window.
A target that begins moving in a later stage receives its initial motion fit
when that stage becomes active. Once initialized, its constant-velocity
hypothesis is reused without periodic FoundationPose tracking or velocity
refitting. Current observations may refresh the visible target mask for
counterfactual editing without updating this motion estimate.

\paragraph{motion prepayment and initial displacement debt.}
\label{app:motion-prepayment-details}
The motion prepayment step modifies executable initial-plan actions while the rest of
the rollout and canonical planning continue. It does not issue another WAM
query. Let $\bm d_0$ be the target displacement already accumulated between
the planning reference captured in $O_0$ and the target position used to
initialize execution, expressed as current target minus planning target.
This \emph{initial displacement debt} is distinct from future displacement
after execution starts. Write $\bm u=\Delta t\,\widehat{\bm v}_{\mathrm{boot}}$
for velocity in meters per policy step. For zero-based action index $i$,
the bounded prepayment used with estimated motion adds
\begin{equation}
 \bm\delta_i=\alpha_i(\bm d_0+e\bm u),\qquad
 \alpha_i=\operatorname{clip}\!\left(\frac{i-s+1}{e-s},0,1\right),
 \quad s=15,\quad e=32.
 \label{eq:appendix-motion-prepayment}
\end{equation}
Thus, actions with $i<15$ are unchanged, and the complete displacement is
reached at the end of action $i=31$, the 32-step boundary. The same
translation is added to the participating end-effectors' reference positions;
reference orientations and gripper commands are preserved. Trajectory inverse
kinematics converts the corrected Cartesian trajectory back to joint commands.
The first chunk is followed by a ten-step terminal continuation: the corrected
terminal EE position advances by $\ell\bm u$ for $\ell=1,\ldots,10$, with
terminal orientation and gripper commands held fixed. This supplies actions
while the canonical replacement is being prepared.

The correction must pass the Cartesian and joint checks in
Table~\ref{tab:execution-controller-settings}. Moreover, the prepared
correction must be ready no later than policy step 15. If it is infeasible or
becomes ready after this activation boundary, the controller skips prepayment
and retains the unmodified initial reference. Bootstrap completion alone does
not guarantee activation: the corrected trajectory must also be prepared in
time. This deadline affects execution-side compensation, not the definition
of the predicted interaction time. The same prepayment ramp and terminal continuation apply to the reported
oracle-velocity evaluations, with the velocity estimate replaced by ground truth.

\paragraph{Ablation: initial-plan warping only.}
\label{app:initial-plan-warping-only}
To test displacement compensation without plan regeneration, we evaluate
FastWAM on the same 192 T2 episodes as the main experiment. Using the
bootstrap motion estimates, we warp the initial grasp-and-release trajectory
toward the predicted object and placement-target positions, preserving its
orientations, gripper commands, and action timing. This extends motion
prepayment across the initial plan, without canonical or current-state
planning; trajectory feasibility checks and the episode budget are retained.
Success decreases from 80.7\% (155/192) with full DPP to 69.8\% (134/192),
counting all episodes, including six in which warping is not activated.
These results suggest that displacement compensation alone is insufficient:
regenerating plans for the predicted target configuration also makes useful
information in the skill prior---including position-dependent approach
motions and interaction behavior---available to execution.

\subsection{canonical planning}
\label{app:wam-full-planning}
\label{sec:canonical-planning}

The canonical planning stage converts the interaction hypothesis into a form directly consumable by the pretrained planner. For the first subtask, DPP synthesizes a canonical observation by preserving the initial robot context and relocating only the target to the predicted interaction position:
\begin{equation}
\begin{aligned}
\widetilde O^{\mathrm{can}}
&= \operatorname{Edit}(O_0,\widehat{\bm x}(\tau)), \\
\widetilde C(S_0,\widehat{\bm x}(\tau))
&= (\widetilde O^{\mathrm{can}},R_0,I).
\end{aligned}
\label{eq:method-canonical-query}
\end{equation}

\begin{figure*}[t]
    \centering
    \includegraphics[width=0.96\textwidth]{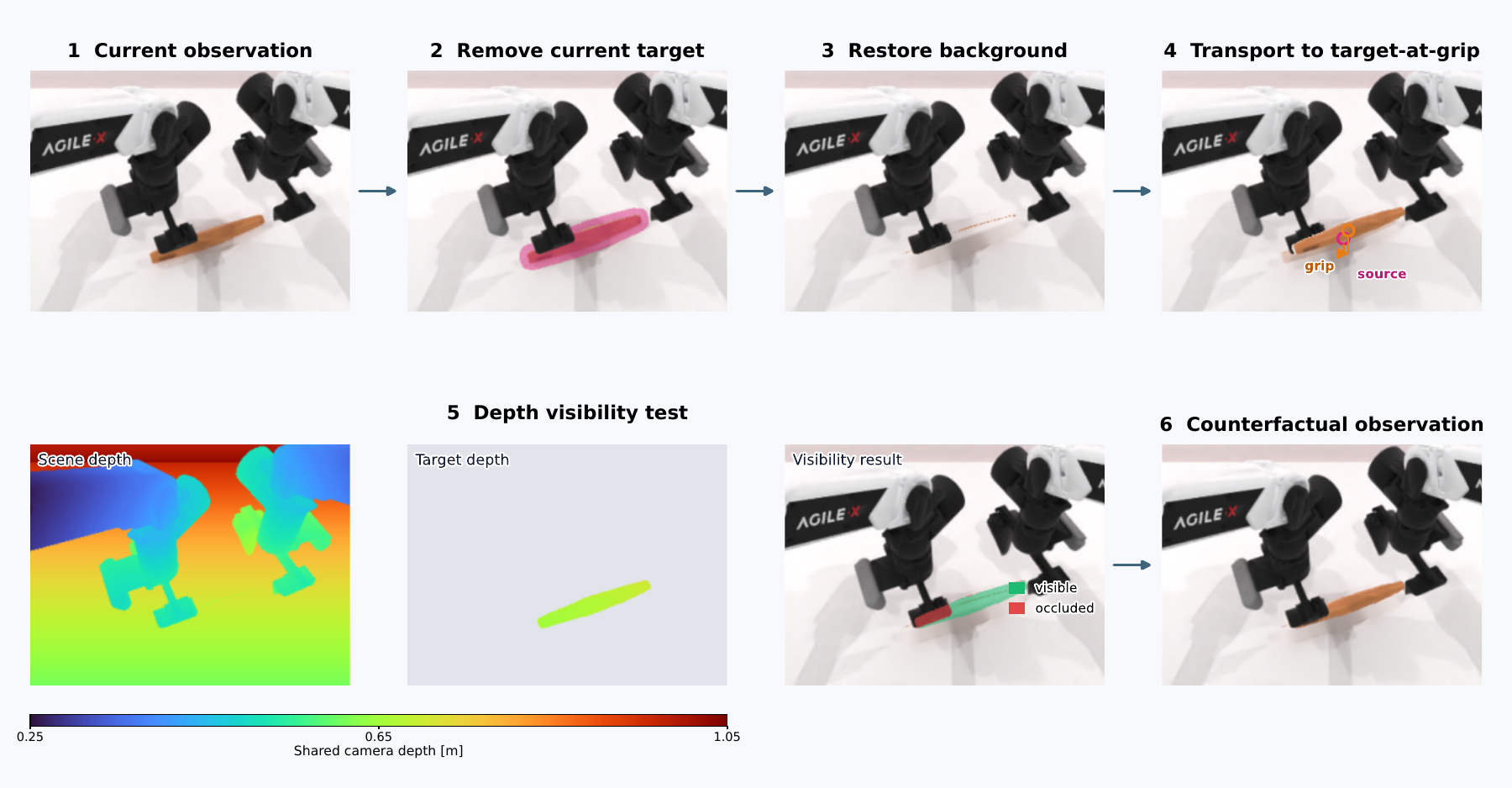}
    \caption{Depth-aware counterfactual observation synthesis with RT-rendered, OIDN-denoised observations. Editing starts from the selected observation. Source-target removal is followed by RGB inpainting and depth restoration using aligned background memory where available. Target reprojection uses scene depth to resolve visibility while preserving the selected robot context.}
    \label{fig:counterfactual-target-inpainting}
\end{figure*}

As in the main text, $\operatorname{Edit}(O,\bm x')$ represents the
designated target at $\bm x'$ while preserving the selected robot and
scene context. Editing starts from the selected observation and removes
the visible source-target region while protecting the robot and manipulated
objects. RGB in the removed region is reconstructed by inpainting from the
surrounding image, while aligned background memory supplies depth where
available. The target RGB--D geometry is then reprojected to
$\widehat{\bm x}(\tau)$, with a z-buffer and scene depth resolving visibility.
Fixed end-effector-to-camera transforms from the robot description map the
target geometry into the wrist views.
Figure~\ref{fig:counterfactual-target-inpainting} illustrates the editing stages.

\begin{figure*}[t]
    \centering
    \includegraphics[width=0.96\textwidth]{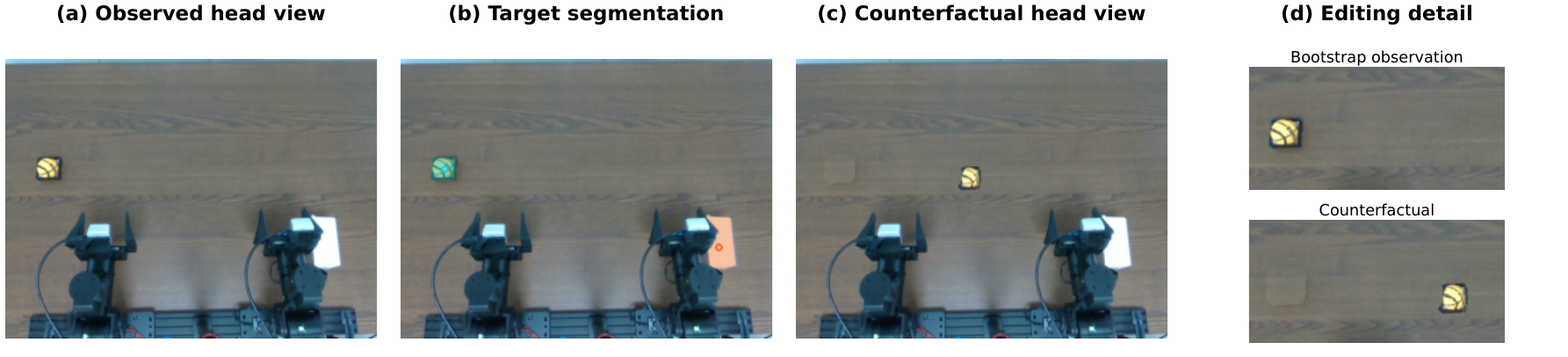}
    \caption{\textbf{Real-robot target detection and counterfactual observation synthesis.}
    A recorded trial shows (a) the bootstrap head observation, (b) its target-segmentation
    overlay, and (c) the synthesized head observation supplied to the first canonical-planning
    request. (d) Matched crops compare the bootstrap and synthesized target regions.
    All panels use recorded pipeline outputs; editing artifacts are retained.}
    \label{fig:real-counterfactual-observation}
\end{figure*}

Figure~\ref{fig:real-counterfactual-observation} shows target segmentation
and the synthesized canonical observation from a real-robot trial, complementing
the simulation editing stages above.

These expressions describe one target; stages involving multiple targets
retain a separate interaction-position hypothesis for each relevant object.
For a static destination that requires no target relocation, the edit is the
identity operation and the observed scene is used directly.

The canonical plan is
\begin{equation}
\zeta^{\mathrm{can}}=\zeta(\widetilde C(S_0,\widehat{\bm x}(\tau))).
\label{eq:method-canonical-plan}
\end{equation}
\paragraph{Rollout budget and termination.}
\label{app:rollout-termination}
Each planning query generates action chunks up to a configured rollout limit,
separate from the episode's execution budget. Generation may terminate early
once the required interaction events and any required continuation have been
predicted. The final generated chunk is retained rather than truncated at the
exact event action. The rollout limit is configured for each task to cover
its required interaction sequence. For example, in Pick-and-Place, detecting
a grasp does not terminate a continuation that must also predict the later
release. A candidate without the
required event cannot supply a valid interaction-aligned replacement;
reaching the rollout limit does not itself constitute an interaction.

In Pick-and-Place, if the planned trajectory is exhausted before task
completion and no planning request is pending, the policy generates additional
action chunks from the current observation to finish the remaining motion.
This continuation can include completing release and retreat, rather than
waiting for an ongoing CP or CSP request.

\subsection{plan acceptance and Activation}
\label{app:plan-activation}
\label{sec:plan-activation}
The canonical plan activation timestamp $t_{\mathrm{can,act}}$ is distinct
from $t_{\mathrm{exec}}$, the time of the first executed action. Activation
does not redefine the frozen predicted interaction time
$\tau$.

\paragraph{Choice of interaction time.}
The skill prior provides a task-dependent estimate of interaction timing,
not a guarantee of the optimal time. An empirically calibrated timing prior
could yield higher success by accounting for the learned manipulation support
and the deployment hardware and environment. In the absence of such
calibration, detecting the event that semantically marks the intended subtask
interaction in the predicted rollout provides a practical default grounded
in the policy's learned behavior, without collecting additional timing-tuning
experience. DPP uses this event-based estimate rather than assuming that its
timing is optimal.

A generated plan does not immediately replace the current action queue.
Acceptance requires a usable finite trajectory, an identified interaction
event, sufficient remaining actions and time, and a feasible connection.
The following rules make explicit the state and timing checks behind the
main-text description.

\paragraph{Predicting the activation state.}
Let $k$ be the current execution step and $\overline T_{\mathrm{bridge}}$
the exponential moving average of transition computation time in seconds.
The controller reserves a lead
\begin{equation}
 L=\max\!\left(2,\left\lceil\overline T_{\mathrm{bridge}}/\Delta t\right\rceil+2\right),
 \qquad k_a=k+L.
\end{equation}
It predicts the robot state at $k_a$ from the already committed action queue
and prepares the connection from that state. Thus, ``current state'' in the
activation description accounts for actions that continue during bridge
computation; it is not the unadvanced request-time joint configuration.
Transition computation estimates are maintained by transition class.

\paragraph{Entry-point selection.}
For base canonical planning, let $g$ be the predicted interaction step in the
candidate rollout, $K$ the frozen interaction step on the execution clock,
and $k_a$ the planned activation step. The zero-based entry index is selected
by temporal alignment alone:
\begin{equation}
 j^*=g-(K-k_a).
 \label{eq:appendix-entry-phase}
\end{equation}
Here, $g$ counts actions through the predicted interaction, so the native
suffix beginning at $j^*$ reaches that event after $K-k_a$ actions.
An entry index outside the available candidate trajectory is rejected.
The controller retains the native action sequence from $j^*$ onward without
temporal resampling. The separate activation bridge connects this suffix to
the predicted robot state at $k_a$; state mismatch is handled by bridge
feasibility rather than by changing the temporally selected entry index.

\paragraph{Bridge construction, acceptance, and fallback.}
For canonical-plan activation, the bridge cross-fades the active and candidate
EE trajectories using the
smoothstep weight $w(r)=3r^2-2r^3$, $r\in[0,1]$; rotations use spherical
linear interpolation with the same blend weight. The nominal procedure
searches for the shortest feasible window between two steps and
$\min(15,K-c-s_{\mathrm{settle}}-k_a)$, where the contact lead $c=0$ and
settling margin $s_{\mathrm{settle}}=2$. It tests Cartesian motion limits and
solves trajectory IK subject to joint-motion checks. Table~\ref{tab:execution-controller-settings}
gives the numerical settings. If insufficient time or queued actions remain,
a valid bridge cannot be activated.

Canonical activation includes a fallback after nominal attempts fail. It uses
the longest available window up to 15 steps and bypasses the Cartesian-limit
rejection, while retaining finite-action and joint-motion checks. Therefore,
the Cartesian limits describe nominal acceptance, not an unconditional bound
on every executed canonical bridge. ``Mandatory'' canonical activation denotes
this additional fallback attempt; it does not guarantee feasibility. A fallback
that also fails the joint checks does not replace the active queue. An accepted
connection installs the bridge and the remaining canonical suffix together.
CSP candidates remain conditional replacements and do not inherit an
unconditional canonical-activation guarantee.

\begin{table}[t]
\centering
\small
\caption{Example controller settings from the main Task~1 constant-velocity
experiments. Steps denote policy steps; prepayment boundaries are relative to the start of the
corresponding planning phase. Cartesian limits apply to prepayment and nominal
Cartesian bridges; the base canonical-to-current joint-space connection uses joint-motion
checks without Cartesian-limit evaluation or trajectory IK. The canonical
fallback retains joint checks but bypasses Cartesian rejection. These are
controller settings, not certified physical safety bounds.}
\label{tab:execution-controller-settings}
\begin{tabular}{@{}lr@{}}
\toprule
Quantity & Setting \\
\midrule
Prepayment start / end boundary & 15 / 32 steps \\
Terminal continuation & 10 steps \\
Added EE speed: prepayment / nominal Cartesian bridge & $0.10 / 0.20\,\mathrm{m/s}$ \\
Total EE speed & $0.50\,\mathrm{m/s}$ \\
EE acceleration / jerk & $5\,\mathrm{m/s^2}$ / $100\,\mathrm{m/s^3}$ \\
Joint speed / acceleration & $2\,\mathrm{rad/s}$ / $20\,\mathrm{rad/s^2}$ \\
Nominal bridge window & 2--15 steps, deadline-limited \\
Contact lead / settling margin & 0 / 2 steps \\
Maximum IK iterations per attempt / nominal attempts & 300 / 2 \\
State / timing scales for state--time phase matching & $0.05\,\mathrm{rad}$ / 4 steps \\
\bottomrule
\end{tabular}
\end{table}

The state and timing scales in Table~\ref{tab:execution-controller-settings}
apply where state--time phase matching is used, including current-state
planning; they do not determine the base canonical-plan entry index, which
is selected by temporal alignment alone. Both IK settings are upper limits.
The nominal attempt limit excludes the additional canonical fallback attempt;
joint-space connections do not invoke trajectory IK.

The prepayment added-speed limit is $0.10\,\mathrm{m/s}$ for T1--T2 and
$0.15\,\mathrm{m/s}$ for the additional-task evaluations, T3--T6.

\subsection{current-state planning and Subtask Transition}
\label{app:current-state-planning}
\label{sec:current-state-planning}

The canonical planning stage uses the subtask's fixed robot reference ($S_0$ for the
first subtask). The current-state planning stage uses fresh head and wrist RGB--D
together with current robot-state input to regenerate the remaining actions
from the state actually reached. Its first request is triggered by a completed
canonical activation bridge, subject to the scheduling rules below. This reanchoring reduces the effect of accumulated rollout error.
In base DPP, the target remains at the same predicted interaction position
within the active subtask:
\begin{equation}
\begin{aligned}
\widetilde O_k^{\mathrm{cur}}
&= \operatorname{Edit}(O_k,\widehat{\bm x}(\tau)), \\
\widetilde C(S_k,\widehat{\bm x}(\tau))
&= (\widetilde O_k^{\mathrm{cur}},R_k,I).
\end{aligned}
\label{eq:method-current-query}
\end{equation}
The resulting current-state plan is
\begin{equation}
\zeta_k^{\mathrm{cur}}=\zeta(\widetilde C(S_k,\widehat{\bm x}(\tau))).
\label{eq:method-current-plan}
\end{equation}
Thus, Canonical and current-state Observations are both counterfactual inputs for a dynamic target. They share the same interaction-conditioned target location but differ in robot context: the former uses the subtask's fixed reference, whereas the latter uses the current state.

The current-state plan remains a candidate until it passes plan acceptance.
Activation resamples only the pre-interaction prefix through joint-position
interpolation, leaving the interaction segment unchanged to preserve the
final approach and gripper closure. For the base canonical-to-current
transition without spatial correction, the controller blends the active and
candidate arm-joint commands using smoothstep weights and selects the shortest
feasible connection window under joint velocity and acceleration limits.
Gripper commands follow the active plan during the blend and the candidate
thereafter. This joint-space connection requires neither trajectory IK nor
Cartesian-limit checks. Base DPP retains the bootstrap motion hypothesis
without spatial correction. The optional nonlinear extension can
update that hypothesis and spatially warp output actions when subsequent
observations reveal substantial deviations, as described in
Appendix~\ref{app:motion-update-options}. A rejected or late candidate leaves
the active plan unchanged within the active stage. After the first request, an eligible target can trigger further current-state
planning once at least 32 policy steps have elapsed since its last request,
regardless of whether the previous candidate was accepted. Requests remain
subject to the active-subtask, remaining-time, and queue-coverage checks below;
they are not restricted to fixed multiples of 32 on the action clock.

\paragraph{CSP request timing and queue coverage.}
Let $\overline\lambda$ be the planner-latency EMA in policy steps and $Q_k$
the number of queued actions. For ordinary transition-aware CSP requests,
the controller requires both $K-k>B$ and $Q_k>B$, where
\begin{equation}
 B=\lceil\overline\lambda\rceil+L+n_{\min}+c+s_{\mathrm{settle}}+2,
 \qquad n_{\min}=2,\quad c=0,\quad s_{\mathrm{settle}}=2.
\end{equation}
Here $L$ uses the appropriate transition-computation estimate. The final two
steps provide additional scheduling margin. The first probe marked due
immediately after a completed bridge uses the relaxed request budget
$B_{\mathrm{first}}=\lceil\overline\lambda\rceil+2$; connection feasibility
is still checked when its result returns. A request passing this scheduling
test is not automatically accepted: the returned plan must retain a valid
interaction event and sufficient horizon, finite commands, and a feasible
activation. Stale or infeasible candidates leave the active suffix in place.

For concurrent interactions, the first completed canonical bridge may trigger
a shared current-state query for multiple targets. Each arm retains its own
interaction deadline and connection checks. A returned candidate for an arm
with a pending canonical bridge is deferred until that bridge has completed;
accepted arm-specific updates are merged into the execution queue.

\paragraph{Rollout reuse and stage transitions.}
Multi-stage tasks compose the same planning operations according to their
interaction structure. When a current-state rollout predicts a subsequent
required interaction, its available prefix can be connected to the active
plan while generation continues. The continuation and its event timing can
then serve as the next stage's initial plan without a duplicate query. At
boundaries configured to start a new planning cycle, DPP instead acquires a
fresh reference and generates a new initial plan.

A new stage uses its selected observation and proprioception as the fixed
canonical reference. Editing preserves the robot and manipulated objects and
relocates only the relevant targets. Already initialized targets retain their
motion hypotheses; targets whose motion begins in the new stage receive their
own initialization window. Static continuation stages use current observations
without moving-target relocation. Within a stage, a missing or rejected
planning result does not itself replace the active suffix; an explicit new-cycle
boundary may reset the queue before initial planning resumes.

For Pick-and-Place, the grasp-stage CSP continuation supplies the release
prior, and the placement reference preserves the grasped object while editing
the remaining target. Handover combines an initial dynamic grasp with
current-observation planning for the subsequent static handover. Stacking
composes sequential pick-and-place cycles: in the evaluated configuration,
each pick-to-place sequence reuses its available continuation, while the
second pick starts a new initial-planning cycle. Its static placement stages
use unmodified observations.

For stacking, frozen predicted interaction times advance the
planner's stage sequence. Simulator events separately govern physical handoff
and task-success evaluation. Together with object- and arm-specific grasp
timing for bottle picking and TCP-based press timing for bell clicking, these
adapters instantiate the common planning procedure through task-specific
interaction predicates and stage composition.
\section{Real-World Experimental Setup}
\label{app:realworld-setup}

\subsection{Robot Platform and Observations}

\begin{figure}[!ht]
    \centering
    \edef\realsetupheight{\the\dimexpr 0.408\linewidth\relax}
    \begin{minipage}[b]{0.40\linewidth}
        \centering
        \includegraphics[height=\realsetupheight,trim=0 0 1050 0,clip]{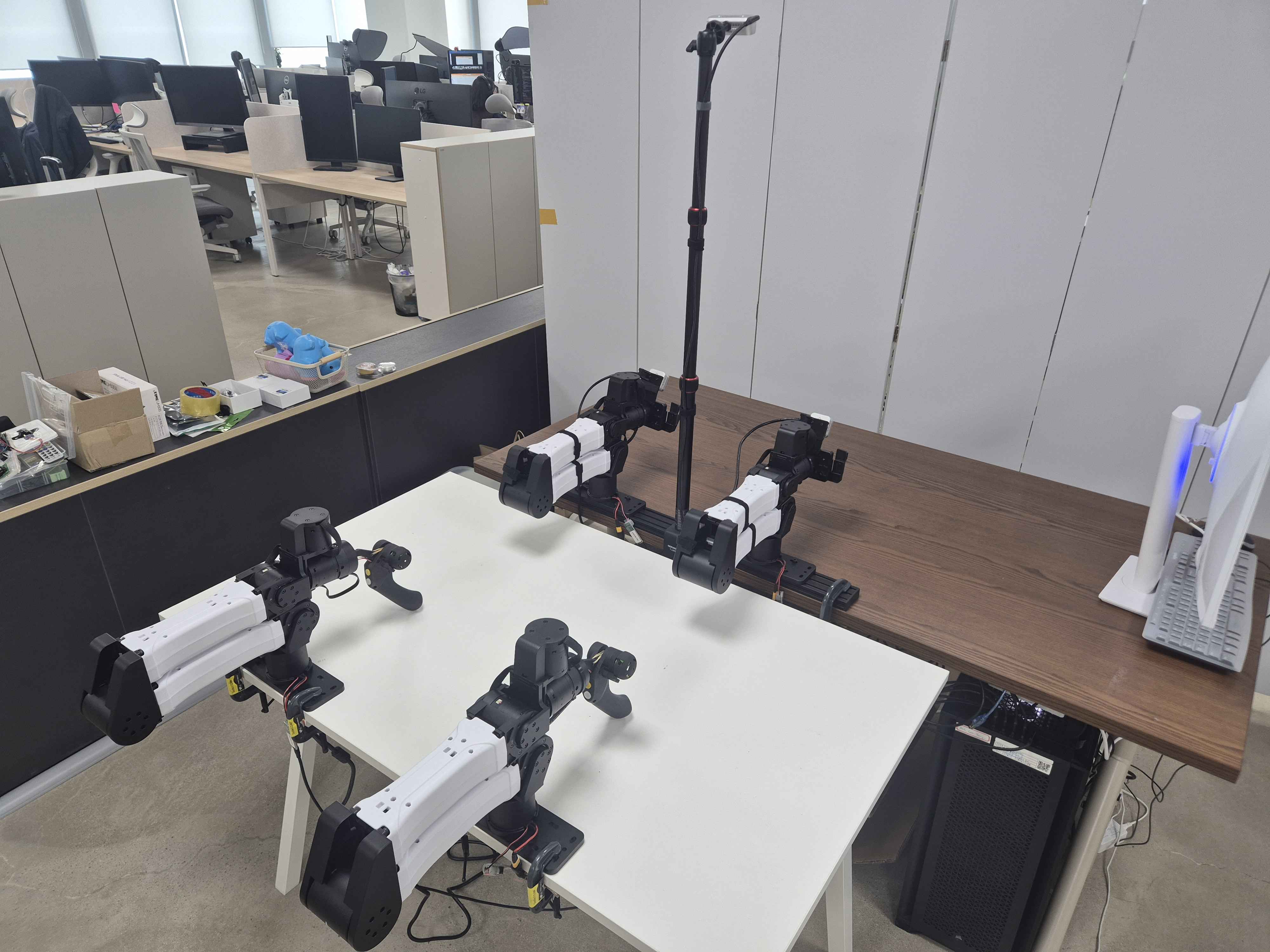}
        \\[3pt] \small (a) Robot platform
    \end{minipage}\hfill
    \begin{minipage}[b]{0.58\linewidth}
        \centering
        \includegraphics[height=\realsetupheight,trim=0 65 0 0,clip]{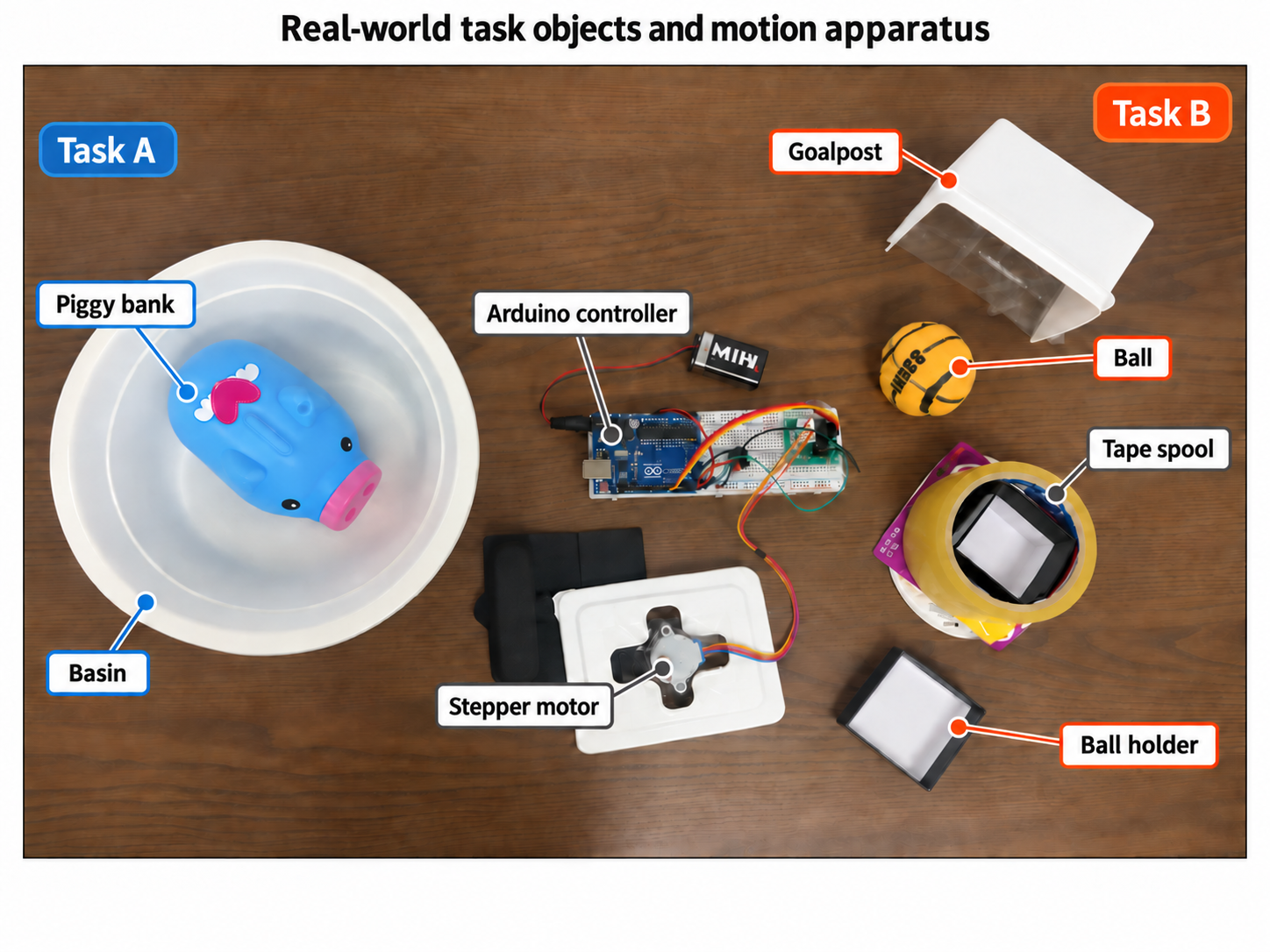}
        \\[3pt] \small (b) Task objects and motion apparatus
    \end{minipage}
    \caption{\textbf{Real-world experimental setup.}
    (a) I2RT YAM leader--follower platform with an elevated head camera and
    wrist-mounted cameras. The follower arms are on the rear rail and the
    leader arms are in the foreground.
    (b) Task objects and the motor-driven apparatus used to generate target
    motion. The objects and apparatus are shown separately for identification.}
    \label{fig:dpp-realworld-setup}
\end{figure}

Our real-world experiments use an I2RT YAM leader--follower bimanual
platform with a tabletop manipulation workspace
(Figure~\ref{fig:dpp-realworld-setup}a). The rig comprises paired leader
arms for teleoperation and follower arms for task execution. An elevated,
fixed head camera overlooks the workspace, and cameras mounted near the
follower grippers provide wrist views. The head view captures the scene
layout, while the wrist views provide local observations around the
end-effectors.

Figure~\ref{fig:dpp-realworld-setup}b shows the task objects and the shared
motion apparatus. Task~A requires grasping a moving piggy bank and placing
it in a basin, while Task~B requires picking up and placing a goalpost so
that the moving ball enters the goal. An Arduino-controlled stepper motor
drives a tape spool to pull the piggy bank in Task~A or the ball holder in
Task~B with a string. The ball remains inside the holder throughout its
motion.

DPP uses timestamped RGB--D observations for target perception and
counterfactual observation synthesis. The WAM planning context combines
visual observations, robot proprioception, and the task instruction.
Depth supports target-geometry reconstruction and visibility-aware
reprojection when placing the target at its predicted interaction position.
The selected robot context is preserved during this editing operation.
Appendix~\ref{app:counterfactual-observation} details the construction, and
Figure~\ref{fig:real-counterfactual-observation} shows recorded perception
and editing outputs from a real-robot trial.

\subsection{Task Objects and Static Demonstrations}

\paragraph{Task A: Pick (Move) \& Place.}
The manipulated object is a moving piggy bank, and the placement destination
is a basin. The robot must intercept and grasp the piggy bank before
placing it in the basin. Success requires completing both the grasp and
the placement; reaching or touching the moving object is insufficient.

\paragraph{Task B: Pick \& Place (Move).}
The robot manipulates a goalpost to intercept a moving ball. Here, the
goalpost is the manipulated object, while the ball supplies the moving
target for placement. Success requires placing the goalpost at a location
that allows the moving ball to enter the goal. This task tests whether the
robot can anticipate the ball's future location when placing the goalpost.

FastWAM is trained only on static demonstrations and remains frozen during
dynamic evaluation. In Task~B, the demonstrations place the goalpost at a
ball-blocking location. DPP and Reactive (Hybrid) use the same policy
trained on these anticipation-oriented static demonstrations, so their
comparison measures how each inference procedure uses the same learned
interaction prior.

\subsection{Target Motion and Perception}

Constant-speed target motion is generated using an industrial tape spool
with a 100\,mm outer diameter, driven by an Arduino-controlled stepper motor.
Operating at 10--14\,rpm produces a target speed of approximately
5.2--7.3\,cm/s. For each fixed experimental setup, we capture one background
reference image with the task objects removed and reuse it across trials.

The perception pipeline uses the background reference to localize the
target, SAM~2 for segmentation, and FoundationPose to track target positions
across timestamped observations. Target motion is fitted to these position
estimates. DPP combines this motion estimate with the
interaction time predicted by the WAM to construct the future target
position used for planning. The segmentation, RGB--D proxy geometry, and
motion-bootstrap procedures are specified in
Appendices~\ref{app:initial-planning-bootstrap}
and~\ref{app:component-validation}.

\subsection{Execution and Evaluation Protocol}

Learned inference runs on a single NVIDIA RTX~4090, hosted by an Intel
Core i7-12700 CPU with 64\,GB of system RAM. Motion bootstrap and visual
planning run asynchronously with nominal 10\,Hz control. The robot executes
available actions while planning continues; an accepted replacement plan
is connected to ongoing execution using the activation procedure
in Appendix~\ref{app:plan-activation}.

Each evaluated method is tested over 30 trials per task, with success
defined by the complete task outcomes above. Reactive replans every
10 action steps. Future-Conditioned uses DPP's motion predictor and
counterfactual observation synthesis with a fixed $+16$-step horizon,
whereas DPP uses the interaction time inferred from the rollout.
Figure~\ref{fig:real-robot-results} reports the real-robot results, including
the shared-policy DPP and Reactive (Hybrid) comparison for Task~B.

\paragraph{Real-world Contact assessment.}
Contact in Task~A is assessed by human evaluation of whether the robot's
gripper contacts the target object during the trial. It is reported separately
from task success, which requires grasping the object and placing it in the
basin. This human-evaluated measure does not use the simulator-specific
gripper thresholds or consecutive-contact sampling rule.

\FloatBarrier

\section{Evaluation Details}
\label{app:dynamic-evaluation}

\paragraph{RoboTwin2.0-Dyn.}
RoboTwin2.0-Dyn provides task-specific criteria for terminating externally
driven object motion and transferring control to physics-driven manipulation.
Rather than applying a single contact criterion uniformly to grasping,
pressing, and placing, it uses handoff conditions that reflect the intended
interaction event in each task. For contact-confirmed grasp events, gripper
closure and target contact are checked jointly, preventing contact by an
open gripper alone from triggering the handoff. These simulator-side handoff
criteria are separate from the planner's predicted interaction times. The
evaluation thus assesses task success under conditions that preserve the
dynamic interaction required by each task, beyond checking the final state alone.

\begin{figure*}[t]
    \centering
    \includegraphics[width=\textwidth]{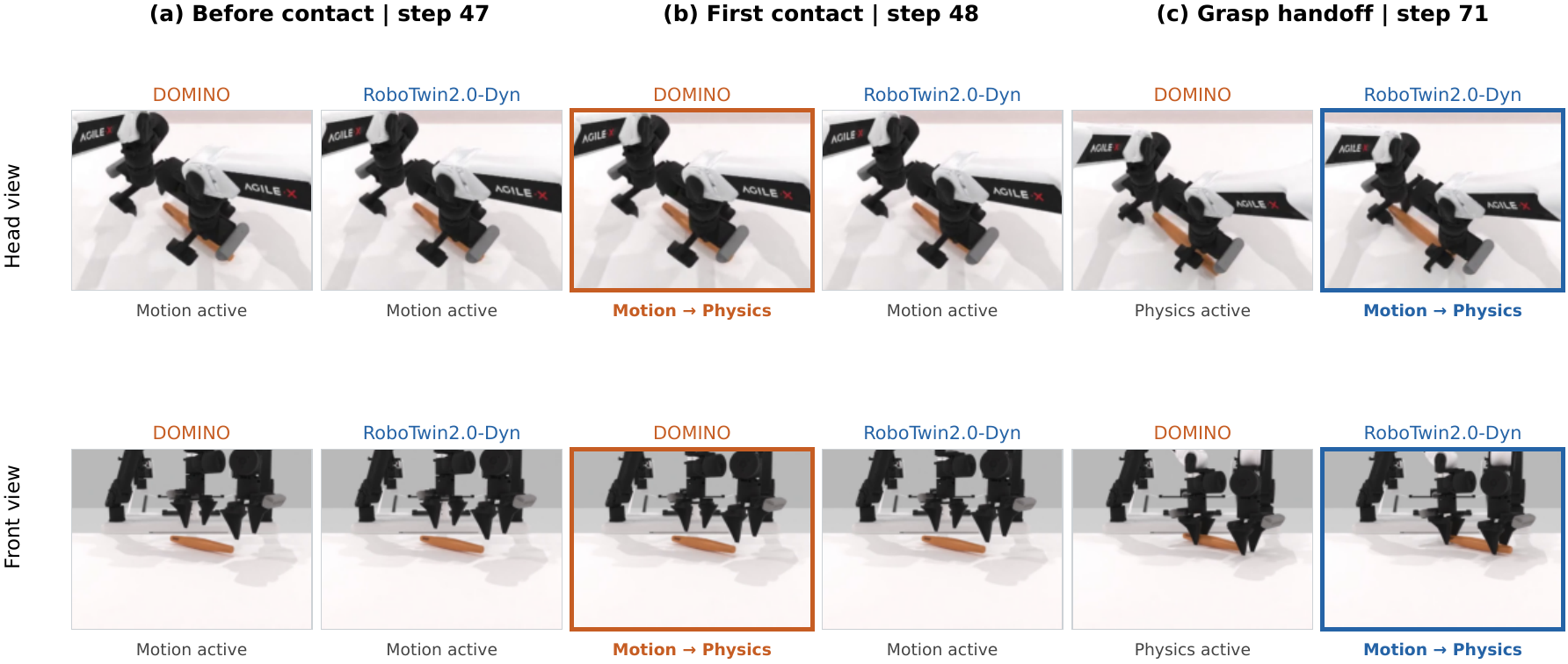}
    \caption{\textbf{Contact-stop versus interaction-aware driver handoff.}
    A roller moves along world $-x$ at 3\,mm per policy step. Each column pair
    compares DOMINO's native driver with the RoboTwin2.0-Dyn T1 driver at the
    same policy step: before contact (47), first contact (48), and the latter's
    grasp-triggered handoff (71). Colored borders mark the corresponding
    motion-to-physics transition frames in both conditions. Head and front views are recorded together
    within each execution; panels show the first recorded observation after
    the indicated action. The two conditions use the same scene, identical
    initial policy observations, and the DPP planner, but are independent
    executions rather than a replay of identical actions. Both satisfy the
    same native task-success criterion.}
    \label{fig:interaction-driver-comparison}
\end{figure*}

\paragraph{Evaluation considerations for DOMINO-trained policies.}
DOMINO constructs dynamic demonstrations by recording a static reference
execution and back-calculating target motion to synchronize the moving object
with the robot's manipulation trajectory~\citep[Sections~2.2 and~D.3]{fang2026towards}.
This construction supports anticipatory interaction at a planned location
and time, providing the training context for DOMINO-trained policies such as
PUMA and DynamicWAM~\citep{fang2026towards,lou2026dynamicwam}.
The DOMINO authors explicitly acknowledge the stop-on-contact simplification
in Section~F.2 and identify richer post-contact dynamics as a direction for
more realistic evaluation~\citep[Section~F.2]{fang2026towards}.
Our evaluation addresses a specific aspect of this acknowledged limitation:
incidental contact before an intended grasp can terminate prescribed motion
and change the interception condition before the grasp is established.
Figure~\ref{fig:interaction-driver-comparison} illustrates this distinction
using the same DOMINO scene and DPP planner: contact-triggered handoff occurs
at step~48, whereas RoboTwin2.0-Dyn's grasp-triggered handoff occurs at
step~71; both executions ultimately succeed.

\begin{wraptable}{r}{0.33\textwidth}
    \centering
    \small
    \caption{\textbf{DOMINO-style evaluation on T1.}}
    \label{tab:domino-contact-stop-diagnostic}
    \vspace{4pt}
    \setlength{\tabcolsep}{4pt}
    \begin{tabular}{@{}lrr@{}}
        \toprule
        Method & Success (\%) & Successes \\
        \midrule
        PUMA & 86.67 & 260/300 \\
        DPP & \textbf{99.67} & \textbf{299/300} \\
        \bottomrule
    \end{tabular}
\end{wraptable}

To assess performance under this early-stopping condition, we apply the
DOMINO-style stop-on-contact rule to the same 300 T1 evaluation scenes.
Because the target can stop before gripper closure establishes a grasp,
we adjust DPP's counterfactual target prediction to anticipate this earlier
stopping event. Specifically, we infer interaction timing using a gripper
threshold of 0.9 and evaluate the target-motion hypothesis five steps earlier
for counterfactual observation synthesis. Execution-time alignment retains
the inferred interaction time, while rollout termination and executed-grasp
detection retain the original 0.2 threshold. With this protocol-specific
adjustment, DPP achieves 99.67\% success (299/300), compared with 86.67\%
(260/300) for PUMA. This result shows that DPP remains effective under
contact-triggered stopping. It also highlights that this evaluation requires
accounting for when the evaluator stops the target before grasping, in
addition to the timing of the grasp itself.

RoboTwin2.0-Dyn refines this handoff beyond simulator contact detection alone
by using task-specific interaction predicates to distinguish incidental
contact from an intended manipulation event. Prescribed motion is maintained
until the corresponding interaction criterion is met, at which point control
passes to physics-driven manipulation. For grasping, this retains the need to
coordinate gripper closure with target arrival, consistent with the
spatiotemporal synchronization used to construct the training demonstrations.
This preserves the interaction structure underlying the training data while
making the motion-to-physics transition sensitive to the task's manipulation
requirements.

\subsection{Main evaluation and table conventions}
\label{app:main-results-details}

Table~\ref{tab:main-dynamic-results} reports simulation results over
300 Task~1 and 192 Task~2 trials per method.
Success and Contact are percentages; Route Completion (RC) is a 0--100 score.
For Task~1, the Contact metric for all methods requires both normalized gripper commands returned by
the robot getters to be below 0.5 and bilateral physical contact with the
object for two consecutive policy steps. Task~2
uses the active arm's grasp contact, as specified below. The Task~1 Contact criterion is applied
post hoc to the recorded trajectories and does not change execution, Success, or RC.

\paragraph{Route completion.}
RC is averaged over all episodes, including failures; a successful episode
receives 100. For a failed episode, let $d_1$ and $d_K$ be the Euclidean
distances from the first and last logged post-action positions to the same
fixed evaluation target. Normalized progress is
\[
    r=\operatorname{clip}_{[0,1]}\!\left(\frac{d_1-d_K}{d_1}\right).
\]
For $d_1<0.01$\,m, the implementation instead sets $r=1$ if
$d_K<0.05$\,m and $r=0.5$ otherwise. In T1, all methods compute $r$ separately
for each TCP and use $100\max(r_{\mathrm L},r_{\mathrm R})$; the fixed target
uses the evalset's final roller XY and the initially logged roller height.
In T2, RC is $100r$ for
the active TCP relative to the final observed placement support. It uses
the episode endpoints without separate grasp/place weights or a
maximum over time. T2 means are episode-weighted over 72 pillbottle,
48 cup, and 72 phone trials.

\paragraph{Contact.}
An episode satisfies Contact when the active arm's gripper value is below
its task threshold and an eligible gripper body of that arm contacts the
grasp object at two consecutive logged policy samples. One eligible
gripper-body contact suffices; successful placement is evaluated separately.
All methods use thresholds 0.9, 0.5, and 0.6 for pillbottle, cup, and
phone, respectively, applied to the normalized gripper command returned by
the robot getter. For baselines, the same predicate is reconstructed from
logged action queues and post-action physical-contact samples.
Dwell is measured in policy samples. T1 RC and both tasks' Contact scores are
recomputed from the saved trajectories using these common definitions.

\paragraph{Method and table conventions.}
Additional Dyn. Training indicates additional policy training for dynamic
tasks beyond the pretrained backbone (O: yes, X: no). +VP denotes the use of video prediction during DPP planning.
Both Future-Conditioned variants use frozen pretrained backbones, refresh every
8 action steps, and condition on a fixed 16-step look-ahead. Oracle uses a
simulator-rendered future observation. DPP bootstrap estimates motion from
observations at steps 0, 2, 4, 6, and 8 before taking policy actions, then
combines this estimate with current RGB-D observations to synthesize the future
input without ground-truth rendering or velocity fallback. These variants are
inference-time proxies for fixed-horizon conditioning, not reproductions of
SIDO's training procedure. Bold values mark the highest value in each metric column.

The FastWAM and ImageWAM observation-refresh baselines assume idealized
zero-latency replanning: robot execution is never paused by inference.
DPP uses the same motion bootstrap, counterfactual observation, plan activation,
and current-state planning controller for both backbones; only the planner
interface differs. The Future-Conditioned variants use neither DPP's
capability-dependent interaction timing nor its full controller.

\subsection{Evaluation protocol}
\label{app:evaluation-protocol}

To account for differences between RoboTwin's predefined evaluation regions
and pretrained manipulation support, we measure
the static-support areas before constructing the dynamic scenes.
The main evaluation regions were selected using FastWAM-based static validation
and then fixed across all compared methods. Starting from
successful seed configurations, we progressively expand the target region,
testing each expansion three times to identify a stable static-support area. The fixed
stable-XY evaluation set uses these empirically validated regions, with the
same scenes and simulator seeds shared across methods, apart from the
archived T2 seed retries described below. Static controls at
the corresponding final target positions distinguish stationary-task
reachability from dynamic execution error.

The Task~1 (bimanual grasp) evaluation set contains 300 moving-target scenes spanning initial
locations, planar directions, and speeds from 1 to 5\,mm per policy step.
Here, one policy step executes one action, distinct from a WAM planning step. The
five speed bins contain 60 scenes each. Every method receives the same scene,
robot initial state, and prescribed constant-velocity trajectory. The evalset
contains neither a grip time nor a per-pair target-stop step.

The Task~2 (pick-and-place) evaluation set contains 192 episodes: 144 co-motion
and 48 counter-motion scenes. Bimanual Grasp requires grasping a moving object,
while Pick-and-Place additionally requires placing the grasped object at a moving
placement target.

The T2 Future-Conditioned (DPP bootstrap) results retain seven archived seed
retries (three FastWAM, four ImageWAM): six after initialization errors and one
after a runtime counterfactual-depth error. Retry seeds are shifted by $10^7$
while preserving episode IDs and prescribed target motion. The six
initialization errors occurred before any policy action; these replacements
retain the motion specification but differ in scene seed from the other
methods. Counting the
runtime-error case as a failure instead changes ImageWAM success from
48.96\% to 48.44\%.

DPP obtains its interaction time $\tau$ from initial planning. For Task~1, the
predicted bilateral gripper sequence triggers the event when both commands
cross the closing threshold of 0.2. The detected rollout offset $j$ is mapped
to $\tau=t_{\mathrm{exec}}+\Delta t\,j$, and the frozen trajectory is evaluated
at that time to obtain $\widehat{\bm x}(\tau)$. This command event
is not a physical-contact label. In the Oracle-free controller, simulator
contact is used only for evaluation: task success follows the official RoboTwin
predicate, while Contact requires both getter-returned gripper commands below 0.5 and
bilateral target contact for two consecutive policy steps. For Task~1, the
official success predicate does not require this sampled contact condition,
so Success can exceed Contact. The matched-prefix physical results in Appendix~\ref{app:target-response-validation} report
official task success.

We compare PUMA and DynamicWAM in their Default execution settings, along with
$\pi_{0.5}$, AHA-WAM, and two Future-Conditioned variants for each WAM backbone.
AHA-WAM uses the official RoboTwin~2.0 checkpoint without additional
dynamic-task training. We retain its default inference configuration:
a 64-action model horizon, 16-action execution chunks, ten denoising steps,
and a six-frame observation history, with a new observation at each chunk.
Both Future-Conditioned variants refresh every 8 action steps with a fixed
16-step look-ahead. Oracle uses simulator-rendered future observations;
DPP bootstrap synthesizes them from estimated motion and current RGB--D.
Both use frozen pretrained backbones without additional dynamic-task training,
as specified in Appendix~\ref{app:main-results-details}.
PUMA and DynamicWAM are evaluated as methods with additional training.
Training-free refresh baselines use Refresh-8 or each backbone's default
cadence (Refresh-24 for FastWAM and Refresh-16 for ImageWAM).
DPP uses FastWAM and ImageWAM without additional policy training.
For Task~1, we report Task Success, bilateral Contact, and dual-arm Route Completion
(RC). This appendix's detailed breakdowns focus on Task~1; the aggregate
Task~2 (pick-and-place) results are reported in Table~\ref{tab:main-dynamic-results}.

\paragraph{Real-robot success criteria.}
Appendix~\ref{app:realworld-setup} specifies the real-robot setup, task objects,
and complete-task success criteria for Tasks~A and~B.

\subsection{Main dynamic-manipulation results}
\label{app:dynamic-results}
\label{app:representative-runtime-timelines}

The complete comparison is reported in Table~\ref{tab:main-dynamic-results} in the main paper.

For Task~2, FastWAM-DPP achieves 80.73\% Success, 96.35\% Contact, and
92.14 RC; ImageWAM-DPP achieves 91.15\% Success, 96.35\% Contact, and
95.99 RC, each over 192 episodes. The DynamicWAM evaluation yields
53.13\% Success, 95.31\% Contact, and 75.89 RC over the same number of episodes.

DPP uses lightweight control observations with a nominal control frequency of 10\,Hz
while planning runs asynchronously. In simulation, target motion follows
wall-clock time during cycle initialization and the action clock after
activation; executed posture-hold actions also advance this clock.
The measured wall-clock action queues below
and real-robot experiments characterize the resulting execution performance.
The DynamicWAM re-evaluation uses accumulated simulator physics time for its
motion-input timestamps, without changing the prescribed target-motion driver.

FastWAM-DPP succeeds in 83.33\% of the 300 scenes and ImageWAM-DPP in 78.67\%.
Observation refresh reaches 10.00--13.00\% for FastWAM and 11.33--12.67\% for
ImageWAM. Future-Conditioned DPP bootstrap reaches 18.33\% with each backbone,
while Future-Conditioned Oracle reaches 27.00\% and 22.67\%, respectively.
The $\pi_{0.5}$ baseline reaches 9.00\%. Among methods evaluated with additional
training, PUMA reaches 50.33\% and DynamicWAM reaches 52.00\%.
Observation access or a short future
condition is therefore
insufficient to recover the support retained by interaction prediction,
canonical observation synthesis, and canonical planning.

\begin{figure}[!htbp]
    \centering
    \includegraphics[width=\textwidth]{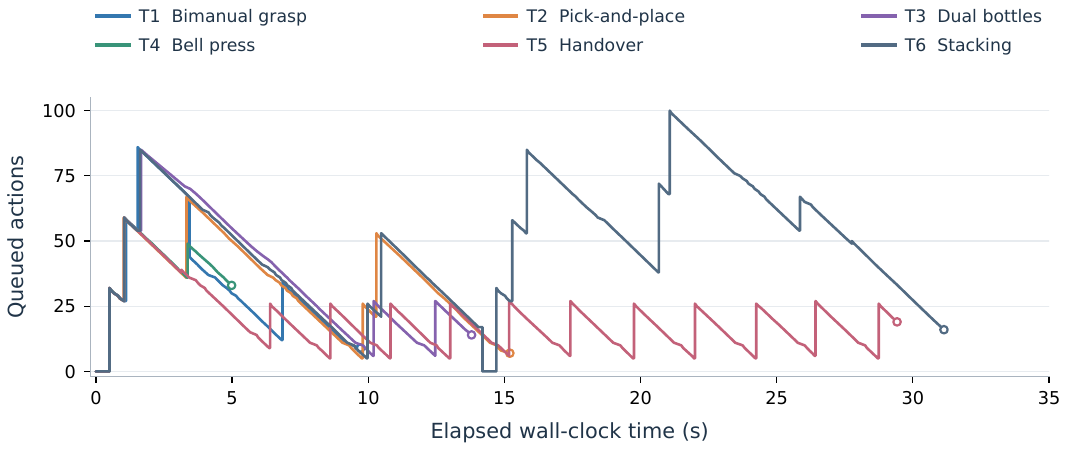}
    \caption{Executable action queues in selected successful T1--T6 episodes.
    A positive queue means actions remain available for execution without
    waiting for further action generation.
    Action-consumption samples are connected linearly for readability, while
    queue replenishment and replacement retain their logged event times.
    Open circles mark the last executed action of each episode. Time includes
    initial and new-cycle preparation within the runtime; unaccepted planning
    candidates are excluded from queue counts.}
    \label{fig:executable-action-queues}
    \label{fig:fastwam-runtime-timelines}
\end{figure}

\paragraph{Asynchronous planning and execution.}
\label{app:control-cycle-statistics}
DPP executes actions at a nominal 10\,Hz while generating plans asynchronously.
With shared initial observations and continuation prefetch enabled, our FastWAM
configuration on an RTX~5090 activates the first action in 0.50\,s after the
initial observation is available and warm-up is complete. A 32-action chunk
takes 0.495\,s to generate on average, including 0.487\,s of backbone inference,
and provides 3.2\,s of execution at 10\,Hz (Table~\ref{tab:planning-latency}).
Subsequent plans are generated while queued actions execute, and continuation
prefetch requests additional actions before the queue is exhausted.
Figure~\ref{fig:executable-action-queues} shows the executable action queues in
six selected successful T1--T6 episodes. No action-queue starvation was recorded
in these runs; zero-queue intervals correspond to initial or new-cycle
preparation. These measurements demonstrate that asynchronous planning supplies
actions ahead of consumption in the tested configuration.

T3 uses Dual Bottle Picking and T2 uses the pillbottle variant. Each trace
uses actual elapsed wall-clock time, without normalization or alignment to
first-action activation. Scene setup, initial observation acquisition, and
model/perception warm-up precede the runtime clock; subsequent observation
acquisition and planning remain timed. These selected examples illustrate
execution behavior rather than aggregate success rates.

\begin{table}[htbp]
\centering
\small
\caption{Measured latency for FastWAM on an RTX~5090, using the six episodes in
Figure~\ref{fig:executable-action-queues} and their 71 generated chunks.
First-action activation is measured from runtime start with the initial
observation available and warm-up complete. Chunk generation includes adapter
processing; backbone inference includes input preparation, denoising,
decoding, and output materialization. These illustrative measurements are
separate from the aggregate task-success evaluation.}
\label{tab:planning-latency}
\label{tab:control-cycle-continuation}
\begin{tabular}{@{}lr@{}}
\toprule
Timing component & Mean latency (s) \\
\midrule
First-action activation & 0.496 \\
32-action chunk generation & 0.495 \\
\quad Backbone inference, included above & 0.487 \\
\bottomrule
\end{tabular}
\end{table}

\FloatBarrier

\subsection{Performance across target velocities}
\label{app:velocitywise-results}

\begin{table*}[t]
    \centering
    \small
    \caption{T1 success by nominal object speed. Each bin contains 60 scenes;
    the All column aggregates 300 scenes and matches Table~\ref{tab:main-dynamic-results}.
    Method definitions follow the main table. Bold marks the highest rate in each column.}
    \label{tab:velocitywise-main-results}
    \resizebox{\textwidth}{!}{%
    \begin{tabular}{@{}llrrrrrr@{}}
        \toprule
        Backbone / Model & Inference Method & \multicolumn{5}{c}{Speed (cm/s)} & All \\
        \cmidrule(lr){3-7}
        & & 1.0--1.8 & 1.8--2.6 & 2.6--3.4 & 3.4--4.2 & 4.2--5.0 & \\
        \midrule
        FastWAM & Refresh-24 (Default) & 25.00 & 13.33 & 8.33 & 3.33 & 0.00 & 10.00 \\
         & Refresh-8 & 26.67 & 15.00 & 13.33 & 6.67 & 3.33 & 13.00 \\
         & Future-Conditioned (DPP bootstrap) & 40.00 & 16.67 & 16.67 & 8.33 & 10.00 & 18.33 \\
         & Future-Conditioned (Oracle) & 56.67 & 23.33 & 20.00 & 15.00 & 20.00 & 27.00 \\
        ImageWAM & Refresh-16 (Default) & 26.67 & 11.67 & 11.67 & 6.67 & 0.00 & 11.33 \\
         & Refresh-8 & 30.00 & 15.00 & 11.67 & 5.00 & 1.67 & 12.67 \\
         & Future-Conditioned (DPP bootstrap) & 33.33 & 21.67 & 15.00 & 8.33 & 13.33 & 18.33 \\
         & Future-Conditioned (Oracle) & 43.33 & 23.33 & 15.00 & 13.33 & 18.33 & 22.67 \\
        \midrule
        $\pi_{0.5}$ & Default & 28.33 & 10.00 & 5.00 & 1.67 & 0.00 & 9.00 \\
        AHA-WAM & Default & 21.67 & 10.00 & 5.00 & 0.00 & 0.00 & 7.33 \\
        PUMA & Default & 71.67 & 65.00 & 45.00 & 33.33 & 36.67 & 50.33 \\
        DynamicWAM & Default & 53.33 & 66.67 & 50.00 & 46.67 & 43.33 & 52.00 \\
        \midrule
        \rowcolor{cyan!7}
        FastWAM + VP & \textbf{DPP} (Ours) & 95.00 & \textbf{81.67} & \textbf{81.67} & \textbf{78.33} & \textbf{80.00} & \textbf{83.33} \\
        \rowcolor{cyan!7}
        ImageWAM + VP & \textbf{DPP} (Ours) & \textbf{96.67} & \textbf{81.67} & 71.67 & 70.00 & 73.33 & 78.67 \\
        \bottomrule
    \end{tabular}%
    }
\end{table*}

At 1.0--1.8\,cm/s, FastWAM-DPP and ImageWAM-DPP reach 95.00\%
and 96.67\% success, respectively. In the 4.2--5.0\,cm/s bin,
they reach 80.00\% and 73.33\%, respectively. DPP is thus not restricted to slow
target motion. Baseline per-speed-bin results are recomputed from the same
300-episode evaluations as their overall rates in
Table~\ref{tab:main-dynamic-results}. Each speed bin contains 60 scenes.

\subsection{Planning-component ablations}
\label{app:component-ablation-diagnostics}

All component ablations use FastWAM on the same 300 T1 scenes.
Removing initial-plan execution retains motion estimation and planning but
holds the robot until CP completes, then executes the native CP trajectory
from its first action. Target motion and the episode budget are unchanged.
Success falls from 83.33\% to 50.33\%, and RC from 94.21 to 83.96.
Mean bilateral gripper closure shifts from 69.1 to 89.1 steps, while the
fraction of CP targets outside the statically evaluated workspace rises
from 27.0\% to 65.0\%. This associates the degradation with delayed
interaction and workspace departure.

Replacing canonical-state conditioning uses current visual and proprioceptive
observations and their corresponding activation procedure at the same request
point. Initial execution, the IP-derived interaction time and predicted target,
and subsequent current-state planning remain enabled. Success is 51.67\%,
Contact is 47.33\%, and RC is 77.96. RC is averaged over all episodes,
including failures.

Removing current-state planning executes the canonical plan without suffix
replacement. Over the same 300 Bimanual Grasp trials, FastWAM-DPP
reaches 83.33\% with current-state planning and 80.00\% without it,
a gain of 3.33 percentage points. The canonical planning stage predicts the interaction
from an earlier robot context; current-state planning uses fresh head/wrist
observations and current proprioception to generate a suffix from the robot
state reached at execution time.

\subsection{Velocity and observation oracle analysis}
\label{app:oracle-analysis}

\begin{table*}[t]
    \centering
    \small
    \caption{T1 oracle ablation, grouped by backbone.
    All settings for both backbones use the same 300 scenes. The Oracle-free
    values match Table~\ref{tab:main-dynamic-results}. Oracle-V and Oracle-Obs
    use the same controller and evaluation protocol with oracle substitutions.
    Contact requires both gripper values below 0.5 and bilateral target
    contact for two consecutive policy steps. RC is averaged over all episodes,
    including failures.}
    \label{tab:dpp-oracle-ablation}
    \begin{tabular}{@{}llrrr@{}}
        \toprule
        Backbone & Setting & Success $\uparrow$ & Contact $\uparrow$ & RC $\uparrow$ \\
        \midrule
        \multirow{3}{*}{FastWAM + VP} & Oracle-free & 83.33\% & 94.67\% & 94.21 \\
        & Oracle-V & 92.33\% & 95.33\% & 97.22 \\
        & Oracle-Obs & 94.00\% & 98.00\% & 97.90 \\
        \midrule
        \multirow{3}{*}{ImageWAM + VP} & Oracle-free & 78.67\% & 93.33\% & 93.43 \\
        & Oracle-V & 90.00\% & 92.67\% & 96.11 \\
        & Oracle-Obs & 93.67\% & 94.67\% & 96.89 \\
        \bottomrule
    \end{tabular}
\end{table*}

Oracle-V replaces only the bootstrap velocity with ground truth while
retaining direct RGB--D synthesis. Oracle-Obs additionally replaces
the synthesized Canonical and current-state Observations with
simulator-rendered RGB--D. These are diagnostic upper bounds and are excluded
from the main comparison.

FastWAM rises from 83.33\% Oracle-free to 92.33\% with Oracle-V and 94.00\% with
Oracle-Obs. ImageWAM rises from 78.67\% to 90.00\% and 93.67\%,
respectively. The current Oracle-free implementation deliberately uses basic,
off-the-shelf SAM~2 and FoundationPose configurations and direct geometric
RGB--D image editing, without task-specific perception training. The oracle
results show that the unchanged DPP controller and WAM backbones retain
substantial headroom. More accurate motion estimation and observation
synthesis can raise performance further, provided that the improved modules
remain within the Canonical and current-state planning latency budgets.

\FloatBarrier

\section{Ablation Details}
\label{app:component-validation}

This appendix validates the deployable perception and systems choices used by
the Oracle-free controller. The goal is not to introduce alternative method
branches, but to separate the behavior of motion bootstrap, counterfactual
observation construction, and runtime scheduling from the end-to-end results
in Appendix~\ref{app:dynamic-evaluation}.

\subsection{Background-referenced target segmentation}

DPP does not require an object category, task actor mask, or task-specific
detector. It compares the initial RGB--D observation with an empty-background
reference acquired under matching scene and camera conditions. RGB difference
and foreground depth evidence define a binary foreground map; morphological
closing and connected-component extraction produce candidate regions. Padded
component boxes prompt SAM~2, and RGB--D plane conditioning removes table
support, shadows, and excess background from the selected masks. In the
multi-object setting, accepted components initialize separate, role-agnostic
instances.
Segmentation produces anonymous object masks; the T3--T6 adapters subsequently
associate them with task roles using declared workspace regions.

This initialization uses the background reference and $O_0$. Candidate and mask validity determine
whether an instance is available; a missing or rejected instance is not
evidence of zero velocity. Background subtraction can detect both static and
moving objects. A static-motion decision instead requires RGB and depth to
remain stable across multiple pre-execution observations. Restricting this
separate gate to pre-execution frames prevents robot motion from being
mistaken for target motion.

\subsection{RGB--D proxy-mesh motion bootstrap}
\label{app:motion-bootstrap}

The Oracle-free estimator uses neither known CAD nor simulator object geometry.
It back-projects RGB--D points inside the initial SAM~2 support to form an
observation-derived proxy mesh, then uses FoundationPose to track its
translation over the buffered bootstrap frames. When pre-motion preparation
is used, initial SAM~2 segmentation, proxy construction, and
FoundationPose object initialization (\texttt{reset\_object}) run during
warm-up. This prepares the tracking sessions; it neither measures velocity
nor establishes that the objects are static.

After the motion epoch begins, the nominal bootstrap capture offsets are
$0,2\Delta t,4\Delta t,6\Delta t,8\Delta t$. The $s_0$ observation is the
single initial reference. Once
the five observations are buffered, the prepared FoundationPose sessions
process them chronologically in an asynchronous worker. Each accepted
instance has its own tracked positions and velocity fit. Actual capture
timestamps, rather than nominal sample indices or tracking completion times,
define the fitting interval; interfaces that store motion-step offsets are
converted consistently using $\Delta t$. A timestamp-aware linear fit
estimates
\begin{equation}
    \widehat{\bm x}_i = \bm b + \widehat{\bm v}_{\mathrm{boot}} s_i + \bm\epsilon_i.
\end{equation}
Here, $\bm b$ is the position intercept and $\bm\epsilon_i$ is the fitting
residual for sample $i$. The stable-XY protocol sets
$\widehat v_{\mathrm{boot},z}=0$ and preserves the
initial target height, preventing depth noise from becoming a spurious
vertical correction. Runtime capture, tracking, fitting, and result polling
remain on the episode timeline even when object initialization was completed
during warm-up. Samples collected after the first action starts remain valid
bootstrap inputs; they are not reused by the pre-execution static gate.

This proxy-mesh pipeline is a minimal real-world-compatible operating point,
not a claim of optimal velocity estimation. The controller accepts the bootstrap motion hypothesis through a modular interface, so a more accurate RGB--D
tracker, learned motion estimator, or multi-view estimator can replace it
without changing canonical planning. The current Oracle-free Task~1 success rates
are 83.33\% for FastWAM and 78.67\% for ImageWAM
(Table~\ref{tab:main-dynamic-results}). The matched-condition oracle
evaluations in Table~\ref{tab:dpp-oracle-ablation} keep the controller and
evaluation protocol fixed while substituting oracle velocity and, subsequently,
oracle observations. They quantify the headroom from improved motion estimation
and observation synthesis. A replacement estimator must still meet the startup
deadline.

\subsection{Frozen bootstrap velocity and optional online refinement}
\label{app:motion-update-options}

The base DPP configuration freezes each target's motion hypothesis after its
initialization window. Targets whose motion begins in a later stage are
initialized separately when that stage becomes active. The nonlinear motion
extension uses continued observations with a declared motion-family prior,
as distinguished below.

Late observations from the current FoundationPose configuration do not have
the same error distribution as bootstrap observations: arm and gripper
occlusion increases, robot geometry enters the depth support, and tracking
bias accumulates over longer intervals. Adding such systematically biased
samples need not improve a linear fit and sometimes shifts the predicted
interaction position away from an already valid canonical plan. Consequently,
base DPP performs no post-bootstrap pose accumulation, online velocity update, or
current-state spatial rebase. The current-state planning stage updates robot context,
while temporal alignment---not spatial warping---aligns the accepted suffix.
A current-frame SAM~2 refresh used for counterfactual image editing updates
the visible removal mask only; it does not restart FoundationPose tracking or
revise the frozen motion hypothesis.

For the nonlinear motion experiments, FoundationPose continues to supply
timestamped poses after bootstrap. A causal fit within the declared motion
family updates the interaction-position estimate, and Late canonical planning
uses the updated hypothesis available immediately before its request. Once
the canonical plan is generated, its interaction-position hypothesis remains
the goal supplied to current-state planning. Subsequent prediction residuals
can instead trigger spatial warping of the current-state plan's output actions;
they do not relocate the target in its planning input. These optional execution
corrections complement temporal alignment and are not enabled in the base
constant-velocity experiments.

\begin{table}[t]
    \centering
    \small
    \caption{Motion-update settings for base DPP and the optional nonlinear
    motion extension. Conditional spatial warping corrects output actions,
    not the target hypothesis in the current-state planning input.}
    \label{tab:motion-update-options}
    \begin{tabular}{@{}lcc@{}}
        \toprule
        Component & Base DPP & Nonlinear extension \\
        \midrule
        Motion-hypothesis fitting & Bootstrap only & Causal, family-specific \\
        Post-bootstrap pose accumulation & Disabled & Enabled \\
        Late canonical planning & Disabled & Enabled \\
        Current-State output spatial warping & Disabled & Conditional \\
        Temporal Alignment & Enabled & Enabled \\
        \bottomrule
    \end{tabular}
\end{table}

\paragraph{Meaning and gates of the nonlinear extension.}
\label{app:nonlinear-gates}
\emph{Late canonical planning} forms the canonical query using a motion
hypothesis updated from causally available observations after bootstrap.
It retains the selected canonical robot context while revising the forecast
of the target at the interaction time. \emph{Spatial warping} instead changes
output EE positions to compensate for the difference between an updated
interaction-position forecast and the incumbent forecast, followed by
trajectory IK. These are distinct operations: changing the planning target
is not the same as correcting an already generated action trajectory.
Neither is enabled in base DPP's constant-velocity configuration.

The reported nonlinear experiments fit the declared motion family causally
from timestamped FoundationPose estimates: constant acceleration,
constant turn rate, or constant velocity with detected position jumps.
New pose requests are spaced by at least four policy steps. Canonical
planning becomes eligible at motion-clock step 24 and uses the latest
available forecast when requested. A short additional prepayment block is enabled
while the delayed canonical plan is prepared. Canonical entry in this
extension uses state--time phase matching before bridge construction.

Current-state planning retains the canonical query's target position in its
input. At completion, the updated forecast can instead correct its output
actions: planar corrections below $0.005\,\mathrm{m}$ are treated as zero, and CSP
candidates requiring corrections above $0.06\,\mathrm{m}$ are rejected.
The same CSP bound applies across the declared motion families. The separate
canonical recovery for a confirmed position jump is not subject to this CSP
correction bound.

This evaluation configuration disables Cartesian-limit rejection for
prepayment and canonical bridges while retaining finite-value checks and
trajectory IK. It uses RoboTwin's MPLib/TOPP joint-limit configuration;
DPP's base numerical joint-speed and joint-acceleration rejection limits are
not applied. These settings are specific to the nonlinear evaluation and
are distinct from the base-controller settings in
Table~\ref{tab:execution-controller-settings}.

\subsection{counterfactual observation validation}
\label{app:counterfactual-observation}

Counterfactual observation synthesis represents the designated target at the
predicted interaction position while preserving the selected robot context.
Editing starts from the selected observation and removes the visible
source-target region while protecting the robot and manipulated objects.
RGB in the removed region is reconstructed by inpainting from the surrounding
image, while aligned background memory supplies depth where available.
The target RGB--D geometry is then reprojected to the predicted interaction
position, with a z-buffer and scene depth resolving visibility.
Fixed camera transforms map the target geometry into the wrist views.

The first canonical observation retains the initial robot context. Current-state
Observations retain fresh head/wrist context and current proprioception while
using the same predicted interaction goal. Subsequent subtasks use their own
fresh canonical robot reference, not the episode-initial state. The Oracle-Obs gains in
Table~\ref{tab:dpp-oracle-ablation} measure the remaining capacity of this
module: replacing direct editing with simulator-rendered RGB--D increases
FastWAM from 92.33\% to 94.00\% and ImageWAM from 90.00\% to 93.67\% after
velocity is controlled.

\FloatBarrier
\subsection{ImageWAM Visual Denoising-Step Reduction}
\label{app:imagewam-denoising-steps}

ImageWAM uses two diffusion loops with different roles.  We retain ten action
denoising steps and reduce only the visual endpoint denoising used between
autoregressive action chunks.  All variants use the same checkpoint,
tensor-core-only \texttt{torch.compile} backend, oracle target velocity, fixed
initial object pose, and matched episode seeds.  The original ten-step visual
sampler is treated as the reference. These latency measurements characterize
this five-pair sampler POC, not a hardware-normalized comparison with the main
benchmark runs.

We evaluated visual step counts $\{10,8,6,5,4,3,2\}$ on five matched
\texttt{grab\_roller} stress pairs spanning left/right diagonal, left/right
lateral, and forward image-plane motion.  Table~\ref{tab:imagewam-denoising}
reports the main latency--quality operating points.  Three visual steps reduce
the combined initial and canonical planning latency by $59.3\%$, while the
initial arm-action RMSE remains $0.0232$ in joint-action units and the mean RGB
difference is $2.16$ on the $[0,255]$ scale.  We therefore use three visual
steps as the default task-level latency--quality Pareto point.

\begin{table}[t]
  \centering
  \small
  \caption{ImageWAM visual denoising-step ablation.  Action denoising remains
  fixed at ten steps.  Latencies are means over five matched episodes; RGB and
  action errors are measured against the ten-step initial rollout.}
  \label{tab:imagewam-denoising}
  \begin{tabular}{lrrrrr}
    \toprule
    Visual & Initial & Canonical & Reduction & RGB MAE & Action RMSE \\
    steps & (s) & (s) & & (0--255 scale) & \\
    \midrule
    10 & 3.840 & 8.321 & -- & 0.00 & 0.0000 \\
     6 & 2.581 & 5.870 & 30.5\% & 0.85 & 0.0042 \\
     3 & 1.726 & 3.226 & 59.3\% & 2.16 & 0.0232 \\
     2 & 1.396 & 2.577 & 67.3\% & 2.79 & 0.0239 \\
    \bottomrule
  \end{tabular}
\end{table}

\begin{figure*}[t]
  \centering
  \includegraphics[width=0.98\textwidth]{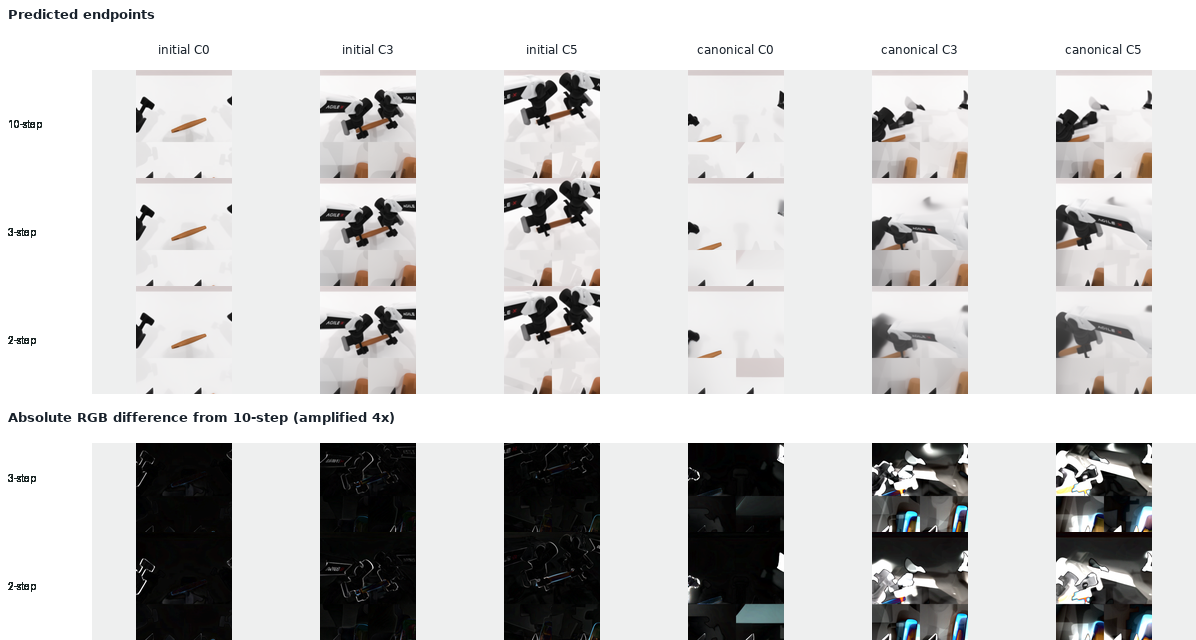}
  \caption{ImageWAM visual denoising boundary on a representative difficult
  pair.  Top rows compare predicted initial and canonical endpoints at fixed
  chunk indices for 10, 3, and 2 visual denoising steps.  Bottom rows show
  absolute RGB differences from the ten-step reference, amplified by
  $4\times$.  Three steps retain the task-relevant geometry with small action
  deviation; the two-step endpoint exhibits the next visible degradation for
  a comparatively small additional latency reduction.}
  \label{fig:imagewam-denoising-cell00}
\end{figure*}

Figure~\ref{fig:imagewam-denoising-cell00} uses a representative difficult
pair to examine visual quality under the same initial scene. In
the initial rollout, three steps retain the ten-step grip event at step 66,
with an RGB MAE of $2.24$ on the 0--255 intensity scale and an arm-action
RMSE of $0.00553$. At two steps, RGB MAE increases to $2.71$ and action
RMSE to $0.00623$, with stronger
residuals around the object and robot boundaries.  The two-step sampler
provides only a further $8.0$ percentage points of aggregate latency reduction
while increasing both visual and action deviation. Accordingly, three steps
preserve a useful margin before this additional visual degradation.

Canonical-rollout errors are interpreted separately from this fixed-index
initial comparison.  Faster samplers complete canonical planning at earlier
environment steps, so canonical differences include both denoising error and
the intended real-time timing change.  The five-pair experiment is therefore a
stress POC rather than a statistical non-inferiority result; broader task
validation remains necessary.

\subsection{Runtime and fallback behavior}
\label{app:runtime-fallback}

For the main simulation experiments, all learned DPP inference components
share one NVIDIA RTX~5090. The simulation host uses an AMD Ryzen~9 9950X CPU
(16 cores, 32 threads) with 64\,GB of system RAM.
The simulator runs separately, with rendering on a second GPU.
Real-robot experiments use a single NVIDIA RTX~4090 for learned inference,
hosted by an Intel Core i7-12700 CPU (12 cores, 20 threads, up to 4.9\,GHz)
with 64\,GB of system RAM. These device assignments distinguish model inference
from simulation infrastructure.

\paragraph{Real-robot target motion and background reference.}
The target-motion apparatus, operating speeds, and reused object-free
background reference are described in Appendix~\ref{app:realworld-setup}.

The initial planning stage and motion bootstrap execute concurrently. The first logical
planning segment is published once it is ready and any required pre-execution
observation collection is complete. The motion prepayment step uses
an available valid motion estimate; object initialization during warm-up is
not itself such an estimate. Once both the interaction timing and bootstrap motion hypothesis are available, they trigger canonical observation synthesis
and canonical planning. FastWAM
returns one native 32-action segment, whereas the ImageWAM adapter combines two
16-action chunks into the same logical segment. Subsequent current-state planning requests follow the minimum 32-policy-step
interval and admission checks described in Appendix~\ref{app:current-state-planning},
rather than being restricted to retries after rejection.

Planner inference and simulator control run in separate processes, and each
backbone and perception service is warmed before episode timing.
An unavailable planning candidate leaves
the active plan intact, and a rejected current-state plan candidate waits for
the next eligible request. Missing perception or planning output never empties
the action queue by itself.

\FloatBarrier

\subsection{Additional Ablations}
\label{app:additional-ablations}

\paragraph{VLA-style control with counterfactual observations.}
We evaluate counterfactual-image-conditioned action generation with VLA-style
receding-horizon execution, without WAM rollout-based planning. We retain
FastWAM as the action generator on the same 300 T1 episodes, isolating the
control strategy while keeping the backbone fixed. We fix the target
hypothesis at interaction step 70, based on the mean interaction time
observed in our T1 evaluations. At each update, the policy receives the
current robot state and a counterfactual observation of the predicted
target position, generates a fixed-length action chunk, and executes its
prefix. This variant achieves 72.33\% success (217/300), compared with
83.33\% (250/300) for full DPP. The policy is conditioned on where the
target is predicted to be, but is not explicitly constrained to complete
the interaction at the corresponding time. The 11.00-percentage-point gap
is consistent with the importance of aligning action execution with the
temporal hypothesis, beyond providing a future-target observation.

\paragraph{Initial-plan warping only.}
We expand on the motion-prepayment ablation in
Appendix~\ref{app:initial-plan-warping-only}. With FastWAM fixed, we evaluate
the same 192 T2 episodes used for full DPP, comprising 144 co-motion and
48 counter-motion episodes. Bootstrap motion estimates spatially warp the
initial grasp-and-release trajectory toward the predicted object and
placement-target positions. Orientations, gripper commands, action timing,
trajectory feasibility checks, and the episode budget are retained, while
canonical and current-state planning are disabled. Success is 69.8\%
(134/192), compared with 80.7\% (155/192) for full DPP, a decrease of
10.9 percentage points. All episodes remain in the denominator, including
six in which warping is not activated. This comparison suggests that
geometric displacement compensation alone does not capture the full benefit
of DPP: regenerating plans also allows the skill prior to supply
position-dependent approach motions and interaction behavior.

\FloatBarrier

\section{Faster Execution and Predictive Adaptation}
\label{app:execution-and-predictive-adaptation}

\subsection{Reaching an Object Before It Moves Versus Where It Will Be}

One intuitive approach to dynamic manipulation is to move the robot faster.
If the robot can reach and grasp an object before it moves substantially,
the positional discrepancy between observation and execution remains small.
This is a valid strategy for mitigating the difficulty of dynamic manipulation,
and faster execution achieved under the same physical constraints is a
legitimate performance advantage.

However, \emph{completing a task before the environment changes substantially
and anticipating those changes are distinct approaches}. Rapidly approaching
an object's currently observed position can be effective when sufficiently
fast robot motion is feasible. In contrast, human--robot collaboration,
collision risks, and grasp stability may require limits on velocity and
acceleration, making it difficult to compensate for object motion through
faster execution alone.

In such settings, moving faster is not the only solution. Anticipating object
motion and coordinating where and when to approach offers an alternative to
rushing the interaction. DPP adopts this perspective: rather than accelerating
robot motion, it focuses on preparing and executing plans that account for a
changing environment.

\subsection{Comparison of Execution Profiles}

To examine policy execution profiles, we compared FastWAM and DynamicWAM on
50 matched evaluation pairs, comprising 25 pairs each from Task~1 and Task~2.
Pairs were selected independently of success. We used a common early execution
window to reduce the influence of differences in episode length.

For each episode, we computed the Euclidean displacement between consecutive
tool-center-point (TCP) positions over policy steps 1--24 and averaged it
across both arms and the 23 transitions. Table~\ref{tab:paired-tcp-displacement}
reports the mean and standard deviation of these episode-level averages.

\begin{table}[t]
    \centering
    \small
    \caption{Mean $\pm$ standard deviation of bilateral TCP displacement over
    policy steps 1--24, measured in mm/action rather than physical velocity.
    Ratios are DynamicWAM/FastWAM ratios of mean displacement. FastWAM uses a
    24-action refresh interval.}
    \label{tab:paired-tcp-displacement}
    \begin{tabular}{@{}lrrrr@{}}
        \toprule
        Group & Pairs & FastWAM & DynamicWAM & Ratio \\
        \midrule
        Task 1 & 25 & $8.61 \pm 0.66$ & $16.87 \pm 0.80$ & 1.96 \\
        Task 2 & 25 & $5.34 \pm 1.63$ & $8.18 \pm 1.49$ & 1.53 \\
        Overall & 50 & $6.97 \pm 2.06$ & $12.53 \pm 4.55$ & 1.80 \\
        \bottomrule
    \end{tabular}
\end{table}

DynamicWAM exhibited greater displacement per action in all 50 matched pairs.
Its overall mean displacement was 1.80 times that of FastWAM, with the
difference particularly pronounced during the earliest actions. These results
highlight substantial differences in how the policies execute their initial
movements.

\subsection{The Direction Pursued by DPP}

Rather than increasing robot displacement per action, DPP focuses on
\emph{preparing a plan for where the object will be}. It estimates an
object-motion hypothesis from observations and constructs a counterfactual
observation reflecting the predicted object position at the anticipated
interaction time. A canonical plan is generated from this observation,
connected to the robot's ongoing execution, and temporally aligned.

Intuitively, the objective is not to reach the currently visible position as
quickly as possible, but to \emph{meet the object where it will be when the
robot arrives}. Object motion is treated as part of the plan rather than
merely an obstacle to overcome through rapid execution. This does not mean
deliberately slowing the robot; it means that accelerating robot motion is
not the central mechanism of the approach.

Faster execution and predictive adaptation are compatible and can be combined.
Their contributions should nevertheless be distinguished. DPP emphasizes
anticipating environmental changes and coordinating execution accordingly,
rather than simply completing the task before those changes become
consequential.

\FloatBarrier

\section{Extended Related Work}
\label{app:extended-related-work}

This appendix expands Section~\ref{sec:related-work} along the four threads DPP ties together: anticipating where a moving target will be, imagining the visual future, reusing skills learned from static demonstrations, and executing while a plan is still being generated. The comparisons follow Sections~\ref{sec:method} and~\ref{sec:evaluation} and concern mechanisms and model interfaces; results reported on different tasks, embodiments, motion distributions, and hardware do not establish an empirical ranking.

\subsection{Dynamic Interaction: Geometry, Motion, and Task Phase}
\label{app:rw-dynamic}

\paragraph{From predicted trajectories to executable interactions.}
Coordinating a robot's motion with a predicted interception state is an old problem. Catching Objects in Flight learns both the object dynamics and the catching motion, and keeps adapting as the trajectory estimate changes~\citep{kim2014catching}; Agile Catching studies constrained whole-body model-predictive control and learned policies for rapid interception~\citep{abeyruwan2023agile}. A grasp must also remain feasible when the robot arrives. Dynamic Grasping with Reachability and Motion Awareness selects grasps using both robot reachability and target motion, and Dynamic Grasping with a Learned Meta-Controller learns how far ahead to predict and how much time to give the motion planner~\citep{akinola2021dynamic,jia2024dynamic}. Anticipating an object's future position and coupling it to robot execution are therefore established ideas. What DPP changes is where the manipulation trajectory comes from: it keeps explicit motion estimation and feasibility checks, but obtains the trajectory from a pretrained WAM. The canonical observation is an interface to a learned skill, not a substitute for geometric reachability or timing.

\paragraph{Learning dynamic support from static demonstrations.}
SIDO factorizes moving-object manipulation into predicting a future object pose and reaching it. Its counterfactual action augmentation displaces the object and the demonstrated action chunk together, so that the trained policy preserves the hand--object relationship~\citep{shin2026static}. DynamicManip generates dynamic demonstrations from a single static one and adapts how often the policy is queried~\citep{liao2026dynamicmanip}. Both carry static experience into dynamic tasks through an augmentation-and-training stage. DPP skips that stage: the model is unchanged, and the intervention happens in the observation supplied at planning time.

\paragraph{Supplying motion and memory.}
PUMA~\citep{fang2026towards} augments policy learning with historical motion and future-state information, and DynamicWAM~\citep{lou2026dynamicwam} adds explicit motion conditioning to a WAM; both appear in our experiments as methods with additional training. RLDX-1 combines temporal and physical sensing within a broader robot model~\citep{kim2026rldx}. TEMPO states the temporal problem most clearly: a single snapshot omits object velocity, and similar-looking scenes at different task phases may call for different actions. It learns to condition on motion summaries and proprioceptive history~\citep{feng2026tempo}; its video encoder is frozen, but the adaptation is not training-free. DPP's matched-relocation probe asks a different question: it fixes the target displacement and varies the robot context, measuring how the response to relocation changes as execution advances. That is a different property from motion perception or phase disambiguation, and our experiments do not show that the two failure mechanisms are independent, nor that canonical querying supplies the temporal information TEMPO learns.

\paragraph{Predicting future features for a frozen action decoder.}
AHEAD is the closest predictive alternative. It trains a latent dynamics module around a frozen VLA and substitutes motion-conditioned future visual tokens into the action decoder, with an uncertainty-based horizon that limits how far it extrapolates~\citep{syed2026intercepting}. DPP instead reads an interaction event off the WAM's own predicted action sequence and queries the skill from a familiar robot context, a reference observation with matching proprioception rather than a state chosen by a prediction horizon. An activation bridge then connects the returned trajectory to the live robot. The two methods share a frozen action backbone; what is specific to DPP is that no predictor or policy is trained for the adaptation.

\paragraph{Anticipating latency versus anticipating interaction.}
F2F-AP predicts sparse object flow, renders it onto the current image, and learns a representation aligned with the observation the robot will see after the estimated sensing, inference, and control delays, so that the policy can run asynchronously~\citep{wei2026f2f}. It is a close precedent for changing a policy's visual context using anticipated motion; the difference is what the horizon means. F2F-AP looks ahead by the system delay. DPP looks ahead to a task event, a grasp or a release, which may lie many inference steps away, and depicts the target at that event in a canonical robot context rather than as a flow overlay on the live observation. DynamicVLA tackles the same deployment problem from the training side, with dynamic data, a compact architecture, and temporally aligned streaming execution~\citep{xie2026dynamicvla}.

\paragraph{Remembering the past and imagining the future.}
MemoryVLA conditions its action expert on perceptual and cognitive context retrieved from a memory bank, and MemoryVLA++ adds predicted future latent states to that retrieved history~\citep{shi2026memoryvla,shi2026memoryvla++}. These methods change what information the policy has available. DPP leaves the policy's inputs as they are and changes the robot--target configuration it is shown.

\subsection{Latency Compensation, Chunk Continuity, and Stopping}
\label{app:rw-execution}

Action chunking is now standard. ACT predicts coordinated chunks and smooths overlapping predictions by temporal ensembling, and Diffusion Policy represents multimodal action sequences and runs them in a receding-horizon loop~\citep{zhao2023learning,chi2025diffusion}. UMI adds that demonstration collection and deployment must respect sensing and actuation timing, including matching inference latency~\citep{chi2024universal}.

Several methods keep execution smooth while a new chunk is computed. Real-Time Chunking generates the continuation asynchronously while honoring the prefix already committed~\citep{black2026real}. Bidirectional decoding chooses among candidate chunks to balance consistency with earlier decisions against responsiveness to new feedback~\citep{liu2025bidirectional}. A2C2 corrects a chunk as new observations arrive, and Delay-Aware Diffusion Policy trains for the delay between observation and execution~\citep{sendai2025leave,liao2025delay}. All of them treat inference, sensing, and execution as separate clocks, as DPP does. DPP's bridge has one extra obligation: the plan it activates was generated from a canonical robot context, so it must be reconciled with the robot's actual configuration, not only with the previous chunk.

How long to trust a chunk is a question of its own. Temporal Action Selection learns to select among cached action predictions~\citep{weng2025temporal}; Adaptive Action Chunking at Inference-time varies commitment with action entropy, and Knowing When to Stop truncates a chunk using an attention-based grounding signal~\citep{liang2026adaptive,xu2026knowing}; Adaptive Action Chunking via Multi-Chunk Q Value Estimation learns values for several horizons in an offline-to-online reinforcement-learning setting~\citep{shin2026adaptive}. These horizons decide how long an existing prediction is executed. DPP's horizon is different in kind: it is the estimated time of a physical interaction, and it decides where the target is placed in the canonical query.

DiscreteRTC completes an asynchronous continuation by discrete diffusion inpainting around committed actions~\citep{wang2026discretertc}, and VLA-Corrector watches for divergence between predicted and observed visual features and triggers corrective replanning~\citep{pan2026vla}. Both manage predictions the policy has already made. DPP changes the context from which the prediction is made in the first place.

\subsection{Visual Prediction as a Planning Model}
\label{app:rw-foresight}

Action-conditioned video prediction gives a learned model of how an intervention changes what the camera sees. Early work on predicting physical interaction and on self-supervised visual planning showed that this model can be learned from robot experience without an analytic scene representation~\citep{finn2016unsupervised,finn2017deep}; temporal skip connections improved how well visual detail persists through the prediction, and Visual Foresight turned the idea into a general planning framework~\citep{ebert2017self,ebert2018visual}. In these systems, the predicted observation is used to score candidate actions against a visual objective. DPP also plans with imagined observations, but uses them differently: it reads an event time off a skill rollout and then poses a new planning query around that event.

Trajectory generation is another route. Diffuser denoises state--action trajectories jointly and conditions on desired outcomes~\citep{janner2022planning}. UniPi generates a task-conditioned video plan and decodes actions from it, Video Language Planning combines visual prediction with language-level reasoning, and UniSim learns a generative simulator of real-world interaction~\citep{du2023learning,du2024video,yang2023learning}. These models differ in whether they predict the consequences of a candidate action, generate a desired future, or provide an environment to learn in, and a predicted video is not by itself an executable plan. In DPP, the WAM's action trajectory carries the manipulation prior, and the execution mechanism absorbs the mismatch between the planning context and the physical one.

Structured prediction widens the interface further. RoboDreamer composes language-conditioned imagination from primitives~\citep{zhou2024robodreamer}, FLIP searches over flow proposals with flow-conditioned video prediction and a value model~\citep{gao2025flip}, and IRASim generates interaction videos conditioned on fine-grained actions for planning and policy evaluation~\citep{zhu2025irasim}.

\subsection{Goal Images, Subgoals, and canonical observation Queries}
\label{app:rw-goals}

Goal-conditioned control exposes the desired outcome as an input instead of baking each objective into a separate policy. Universal Planning Networks learn representations suited to planning toward goals, RIG learns visual goals for reinforcement learning, and LEAP plans over latent subgoals~\citep{srinivas2018universal,nair2018visual,nasiriany2019planning}; Goal-Conditioned Predictors generate the intermediate observations that a long-horizon visual plan passes through~\citep{pertsch2020long}. Generative image models make such targets easy to write down: SuSIE edits the current image into a subgoal for a goal-conditioned policy, DALL-E-Bot generates the goal arrangement of a scene, and Dream2Real links imagined goal states to physical manipulation~\citep{black2024zero,kapelyukh2023dall,kapelyukh2024dream2real}.

A goal image tells a goal-accepting controller where to end up. DPP's edited observation is instead the conditioning input to an existing WAM: it depicts the target at its hypothesized interaction position together with a familiar robot state, so the model treats it as where the skill starts. Editing the conditioning observation therefore changes the model's assumed starting context, not only its endpoint, which is why the returned canonical trajectory needs an explicit connection to live execution.

\subsection{World--Action Modeling and the Cost of Imagination}
\label{app:rw-wams}

Predicting future images alongside actions predates the WAM label. GR-1 pretrains on video and then predicts future frames and robot actions from language, observation history, and robot state, and GR-2 scales the same recipe with broader video pretraining and joint video--action learning~\citep{wu2024unleashing,cheang2024gr}. Generalist action models such as RT-1, RT-2, OpenVLA, GR00T~N1, and $\pi_{0.5}$ supply reusable visuomotor priors from large and diverse data~\citep{brohan2022rt,brohan2023rt,kim2024openvla,bjorck2025gr00t,intelligence2025pi_}. DPP addresses one deployment situation for such a backbone: it already has the static manipulation skill, yet fails when the target moves.

VideoVLA, ViPRA, LingBot-VA, and DreamZero develop the coupling between video prediction and action generation in different ways, and Cosmos~3 extends world modeling across modalities~\citep{shen2026videovla,routray2026vipra,LiL-RSS-26,ye2026world,agarwal2026cosmos}. For DPP, the architectural details matter less than one property: whether the interface can return both future observations and the corresponding actions across a skill rollout. Joint and cascaded designs both qualify, provided their predictions can be rolled out recursively.

Because imagining the future is expensive, several recent WAMs remove it from the deployment path. Fast-WAM drops explicit future imagination from its standard test-time action path, ImageWAM works with image-editing representations, Efficient-WAM lowers the cost of imagination, and Flash-WAM applies modality-aware distillation~\citep{yuan2026fast,zhang2026imagewam,li2026efficientwam,akbari2026flash}. AHA-WAM couples a slower world planner to a faster feedback policy through cached latent context, and can likewise omit explicit future video at deployment~\citep{cai2026aha}. Our experiments run DPP on FastWAM and ImageWAM through adapted interfaces that expose future observations and actions for recursive rollout; this is a requirement of our planner, not a description of these backbones' default inference. DPP uses the rollout to estimate the interaction event while overlapping planning with execution.

\subsection{Data Augmentation, Observation Editing, and Policy Steering}
\label{app:rw-interventions}

Augmentation enlarges what a policy sees during training. GenAug retargets demonstrations to new scenes with generative visual augmentation, ROSIE uses text-to-image models for semantic augmentation at scale, and MimicGen assembles new demonstrations by transforming and composing segments of existing ones~\citep{chen2023genaug,yu2023scaling,mandlekar2023mimicgen}. SIDO is the most direct dynamic-manipulation counterpart, since it ties counterfactual object displacement to the corresponding action change~\citep{shin2026static}. DPP works at the other end of the pipeline: the model stays fixed and its planning input changes at deployment. The two are complementary: augmentation can broaden the backbone's skill coverage before deployment, and DPP's effectiveness is bounded by that coverage.

ReOI is the closest precedent for editing an observation at inference time in world-model planning. It removes unfamiliar distractors before the model predicts action outcomes, then restores them to verify the result~\citep{chen2025reimagination}. DPP shares the goal of making a query look more like the model's experience, but edits the task-relevant target rather than distractors, and pairs it with a canonical robot context so that the model generates the interaction trajectory. VLS intervenes at a different point: a vision--language model constructs rewards that guide a frozen generative policy's action sampling~\citep{liu2026vls}. DPP instead acts through observation conditioning, followed by geometric and temporal checks on execution. The two interventions also need different access: observation editing needs a way to construct the scene, and reward-guided sampling needs access to the generative action process.

DPP's \emph{counterfactual} observation is simply a synthesized planning observation that differs from the live scene. The canonical reference is the initial robot state for the first subtask and a fresh post-interaction reference for later ones. The robot keeps moving while the plan is generated and joins the canonical trajectory through an activation bridge, after which current-state planning uses fresh observations and proprioception while keeping the predicted interaction position. Base DPP holds its bootstrap motion hypothesis fixed and applies no spatial correction to current-state actions; updating the motion estimate and correcting those actions belong to the optional nonlinear-motion extension.

\subsection{Imitation-Learning Distribution Shift and the Scope of Action Support}
\label{app:rw-support}

Imitation learning must cope with the gap between demonstrated states and the states the learned policy actually visits. DAgger closes it by aggregating data under the learner's own distribution, and DART by perturbing the expert so that the demonstrations already cover likely errors~\citep{ross2011reduction,laskey2017dart}. Both supply recovery experience through training. DPP assumes no such additional adaptation is available: the backbone is reused as is, and planning tries to reach an existing skill through a more familiar pairing of robot state and target placement.

The evidence in Section~\ref{sec:target-response-collapse} should be read with the same care. The matched-relocation probe measures a loss of responsiveness to target relocation across robot execution contexts. Uneven demonstration coverage of robot--target combinations is a plausible explanation, but the probe measures generated behavior, not training-distribution density. It motivates canonical querying within the evaluated static support; it does not prove that support mismatch is the only cause of failure. Missing temporal information, prediction error, delayed actuation, and insufficient static competence can all contribute~\citep{feng2026tempo,liao2025delay}.

\subsection{Benchmarks and the Limits of Cross-Paper Comparisons}

LIBERO studies knowledge transfer across manipulation tasks, and RoboTwin builds dual-arm evaluation on generative digital twins~\citep{liu2023libero,mu2025robotwin}; RoboTwin~2.0 adds scalable task synthesis and structured domain randomization, and provides the tasks we use~\citep{chen2025robotwin}. Making these tasks dynamic requires specifying target motion and interaction timing as well. UMI offers a complementary view in which the demonstration interface and its timing calibration are what make dynamic skills learnable~\citep{chi2024universal}. Our static-support validation has a narrower purpose: it separates failures of dynamic execution from the absence of the underlying static skill, so that what we measure is reuse of a skill the backbone already has under declared motion conditions.

The comparisons we report should be read accordingly. The Future-Conditioned baselines run the same frozen backbones with a fixed 16-step look-ahead; they isolate fixed-horizon conditioning and are not reproductions of SIDO's augmentation and training pipeline. PUMA and DynamicWAM are the trained dynamic comparators. AHEAD, TEMPO, F2F-AP, ReOI, and VLS are compared here at the level of mechanism only, and nothing in our experiments ranks DPP against them. The real-robot goalpost task also depends on static demonstrations that already encode the intended interception geometry.

\subsection{Comparison of Closely Related Adaptation Mechanisms}

Table~\ref{tab:rw-mechanisms} summarizes the intervention each of the closest approaches makes and how it relates to DPP. The rows describe mechanisms, not a matched experiment; combining any of them with DPP would require its own controlled evaluation.

\begin{table}[ht]
\centering
\small
\setlength{\tabcolsep}{4pt}
\renewcommand{\arraystretch}{1.13}
\begin{tabular}{p{0.18\linewidth}p{0.35\linewidth}p{0.40\linewidth}}
\toprule
Approach & Intervention & Relation to DPP \\
\midrule
SIDO~\citep{shin2026static} & Augment object poses and the associated action chunks during training. & Shares future-pose conditioning; DPP changes the frozen model's planning observation instead. \\
AHEAD~\citep{syed2026intercepting} & Train a latent predictor of future visual tokens for a frozen VLA. & Shares anticipation; DPP uses the WAM's own rollout and a canonical robot context. \\
TEMPO~\citep{feng2026tempo} & Learn conditioning on motion summaries and proprioceptive history. & Addresses missing temporal information; DPP probes responsiveness across robot contexts. \\
F2F-AP~\citep{wei2026f2f} & Predict a flow-based observation representation one system delay ahead. & Shares motion anticipation; DPP places the target at a predicted task event. \\
RTC~\citep{black2026real} & Generate continuations asynchronously around committed actions. & Addresses chunk continuity; DPP must also connect a counterfactual trajectory to the live state. \\
AHA-WAM~\citep{cai2026aha} & Couple a slower world planner with a faster feedback action model. & Shares asynchronous world-model control; DPP adds canonical querying at the interaction target. \\
ReOI~\citep{chen2025reimagination} & Remove distractors before prediction and restore them for verification. & Shares test-time observation editing; DPP relocates the task target to generate the skill. \\
VLS~\citep{liu2026vls} & Guide a frozen policy's sampler with rewards from a vision--language model. & Adapts an existing policy through action sampling rather than through the observation. \\
DPP (ours) & Predict interaction time, query a canonical observation, and bridge to execution. & Reuses static skill support under predictable target motion, with no additional adaptation training. \\
\bottomrule
\end{tabular}
\caption{Mechanistic comparison with the closest related work. A frozen backbone does not make a method training-free: auxiliary predictors or conditioning modules may still be trained.}
\label{tab:rw-mechanisms}
\end{table}

\end{document}